\documentclass[runningheads]{llncs}
\usepackage[T1]{fontenc}
\usepackage{listings}
\usepackage{xcolor}
\usepackage{mdframed}
\usepackage{subcaption}
\usepackage{graphicx}
\usepackage{amsmath}
\usepackage{amssymb}
\usepackage{booktabs}
\usepackage{stfloats}
\usepackage{placeins}
\usepackage{float}
\usepackage{array}
\usepackage{enumitem}
\usepackage{tabularx}
\usepackage{makecell} 
\usepackage{graphicx}
\usepackage{siunitx} 
\usepackage[numbers]{natbib}

\usepackage{color}
\usepackage{hyperref}
\hypersetup{
    colorlinks=true,  
    linkcolor=black,  
    citecolor=black,  
    filecolor=black,  
    urlcolor=black,    
    anchorcolor=black 
}

\usepackage[nameinlink, noabbrev]{cleveref}

\definecolor{codeblue}{rgb}{0.122, 0.435, 0.698} 
\definecolor{codelightbg}{rgb}{0.95, 0.95, 0.97} 

\newmdenv[
  backgroundcolor=codelightbg,
  linecolor=codeblue,
  outerlinewidth=8pt,
  roundcorner=10pt,
  innertopmargin=10pt,
  innerbottommargin=10pt,
  leftmargin=10pt,
  rightmargin=10pt,
  skipabove=\baselineskip,
  skipbelow=\baselineskip
]{codebox}

\Crefname{figure}{Fig.}{Figs.}

\setlist[itemize]{noitemsep, topsep=0pt, partopsep=0pt}
\setlist[enumerate]{noitemsep, topsep=0pt, partopsep=0pt}



\begin{document}
\title{\texorpdfstring{Iris-SAM: Iris Segmentation Using a \\Foundation Model}{Iris-SAM: Iris Segmentation Using a Foundation Model}}
\titlerunning{Iris-SAM: Iris Segmentation Using a Foundation Model}
%
\author{Parisa Farmanifard\inst{1}
\and
Arun Ross\inst{1}}
%
\authorrunning{P. Farmanifard and A. Ross}
%
\institute{Michigan State University, East Lansing, MI 48824, USA \\
\email{\{farmanif,rossarun\}@msu.edu}}
\maketitle              
\begin{abstract}
Iris segmentation is a critical component of an iris biometric system and it involves extracting the annular iris region from an ocular image. In this work, we develop a pixel-level iris segmentation model from a foundation model, viz., Segment Anything Model (SAM), that has been successfully used for segmenting arbitrary objects. The primary contribution of this work lies in the integration of different loss functions during the fine-tuning of SAM on ocular images. In particular, the importance of Focal Loss is borne out in the fine-tuning process since it strategically addresses the class imbalance problem (i.e., iris versus non-iris pixels). Experiments on ND-IRIS-0405, CASIA-Iris-Interval-v3, and IIT-Delhi-Iris datasets convey the efficacy of the trained model for the task of iris segmentation. For instance, on the ND-IRIS-0405 dataset, an average segmentation accuracy of 99.58\% was achieved, compared to the best baseline performance of 89.75\%.

\end{abstract}
\section{Introduction}
\label{sec:Introduction}

\begin{figure}[ht]
  \centering
  \hspace*{-3em} 
  \includegraphics[width=0.75\linewidth]{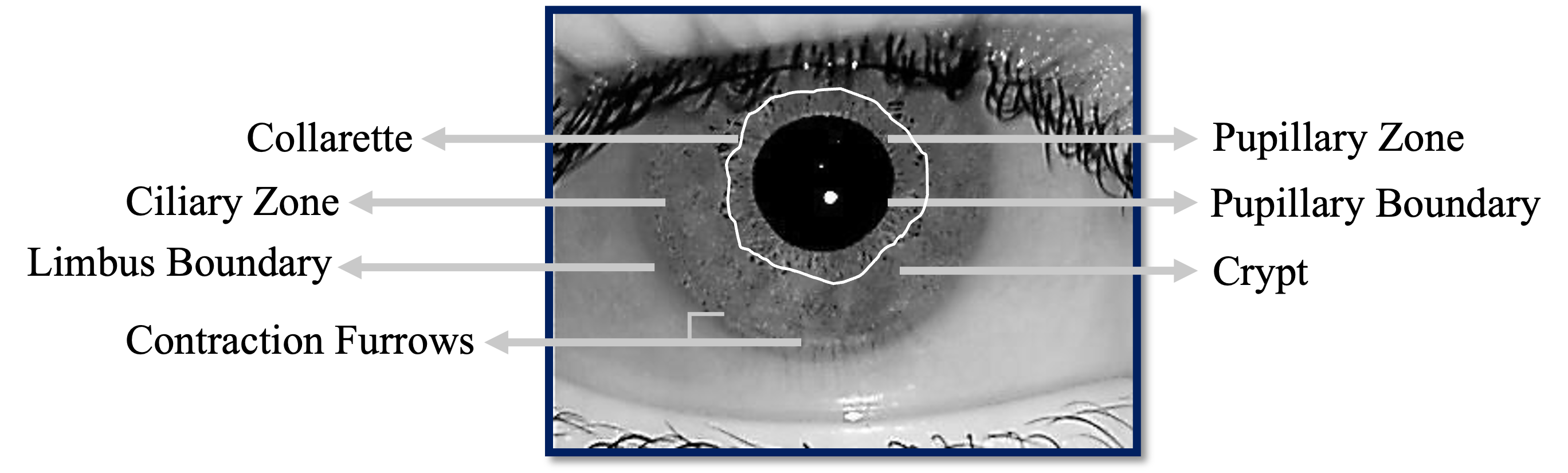}
  \captionsetup{labelsep=period} 
  \captionsetup{labelfont=bf}
  \caption{Iris anatomy in the near-infrared spectrum.}
  \label{fig: anatomy}
\end{figure}

The human iris is a highly intricate structure whose distinctive patterns form the basis of one of the most secure forms of biometric recognition \cite{daugman2009iris,jain201650}. The structure of the iris includes the pupillary zone immediately surrounding the pupil, the ciliary zone comprising the rest of the iris, and the collarette forming a boundary between the two zones. The iris also features contraction furrows, which are circular folds created by the contraction of the pupillary muscles, and crypts, which are small openings or depressions in the iris tissue (\Cref{fig: anatomy}) \cite{10.5555/2161587}.
\vspace{+2mm}

 The process of accurately delineating the iris from the surrounding structures (sclera, pupil, eyelids, eyelashes) is termed iris segmentation \cite{nguyen2017long}. A well-segmented iris results in more reliable feature extraction, which, in turn, enhances the accuracy of recognition systems. Several iris segmentation methods have been developed, ranging from the simple Integro-Differential Operator \cite{daugman1993high} to those based on complex convolutional neural networks \cite{jalilian2017iris}. However, iris segmentation has its own challenges. Variability in illumination, occlusions from eyelashes or eyelids, reflections, and low resolution in distant captures are among the issues complicating the process \cite{bowyer2008image} and can substantially affect the segmentation quality. Additionally, intrinsic challenges such as the dynamic nature of the iris, aging effects, and diseases can further complicate the segmentation task. 
 \newline \indent Traditional segmentation methods have evolved significantly with the integration of deep learning techniques. Fully Convolutional Network (FCN) \cite{long2015fully} and U-Net \cite{RFB15a} have laid the groundwork for advanced segmentation tasks. These models have been instrumental in achieving state-of-the-art performance in various segmentation challenges, demonstrating the power of deep learning in extracting meaningful patterns from complex visual data.
 \newline \indent Further, the recent surge in the capabilities of foundation neural network models offers a promising path for advanced segmentation. The advent of foundation models has revolutionized various domains of research and application. These models, characterized by their vast size, extensive training data, and remarkable generalization abilities, have set new benchmarks in tasks ranging from natural language processing to image recognition \cite{brown2020language, he2016deep}. 
 \newline \indent Among the recent advancements in foundation segmentation models, the Segment Anything Model (SAM) by Meta stands out as a versatile and powerful tool. SAM's architecture, designed to handle a wide range of segmentation tasks, represents a significant leap in the field \cite{kirillov2023segment}. Its ability to segment diverse objects in complex scenes has shown great promise in extending the boundaries of what segmentation models can achieve. However, applying foundation models like SAM to specific domains, such as iris segmentation, presents unique challenges. As explained, the complexity of iris patterns, variations in lighting, occlusions due to eyelashes, varying pupil sizes, the pose of the iris with respect to the camera, the acquisition of the image in the near-infrared spectrum, etc., are significant hurdles in achieving high-precision segmentation \cite{daugman2009iris}. The transfer of knowledge from general segmentation tasks to such a specialized domain requires careful fine-tuning and adaptation of the model. Approaches like transfer learning, where a model trained on a large dataset is fine-tuned on a smaller, domain-specific dataset, have proven effective \cite{pan2009survey}.
 \newline \indent The central inquiry of this study thus emerges: how effective are these foundation models, particularly the Segment Anything Model (SAM), when applied to the task of iris segmentation? By exploring this question, in this paper, we delve into the challenge of iris segmentation, a crucial component for iris recognition and other tasks such as eye tracking, gaze estimation and proximity detection in eye surgery, and investigate how foundation models can be fine-tuned and augmented with specific loss functions \cite{8417976} to achieve enhanced performance.

\section{Background}
\subsection{Foundation Models}
One of the most popular foundation models, OpenAI's GPT-3, was trained on diverse internet text. This vast training allows it to write essays, answer questions, and even generate code with minimal task-specific prompts \cite{brown2020language}. Google's BERT uses bi-directional training on vast text corpora to capture intricate language relationships, setting new performance benchmarks across numerous NLP tasks \cite{devlin2018bert}. In the visual domain, Vision Transformers (ViT) tessellate images into fixed-size patches, linearly embed them, and process them through transformer layers \cite{dosovitskiy2020image}.

While foundation models emerged with a strong emphasis on NLP tasks, their underlying architecture and learning mechanisms have found relevance in computer vision. The ability of these models to capture hierarchical features and understand contexts makes them suitable for visual tasks, where often, object representations and spatial relationships dictate the outcome. For instance, Vision Transformers (ViT) have demonstrated how techniques originally designed for sequence data (like text) can be repurposed for image data, leading to impressive results in image classification \cite{dosovitskiy2020image}. The success of such models suggests potential utility in more intricate tasks like segmentation, where both global and local contexts matter.

\subsection{SAM}
The Segment Anything Model (SAM) \cite{kirillov2023segment} exemplifies the adaptability of foundation models in complex computer vision tasks like segmentation. Its architecture includes an image encoder creating high-dimensional embeddings, a prompt encoder for interpreting user inputs (like bounding boxes) into features, and a mask decoder that merges these to predict segmentations. SAM's vision encoder, a key component, processes images to generate detailed feature maps, crucial for capturing contextual, localizing, and pattern-recognition information across large image areas. The mask decoder then translates these features into precise pixel-level decisions for segmentation, using up-sampling layers to refine the output. SAM's ability to acquire a broad knowledge base from varied training scenarios enables it to adapt to new segmentation tasks with minimal additional training, bridging foundation model capabilities with the specific needs of segmentation accuracy.

\subsection{Iris Segmentation}
Over the years, iris segmentation methods have evolved from traditional image processing techniques to sophisticated machine learning algorithms. The seminal work of Daugman utilized integro-differential operators to detect the iris' circular patterns by maximizing contour sharpness \cite{daugman2009iris}. This approach, however, has its limitations with non-circular and partially occluded irides. Subsequent advances have introduced active contour models \cite{shah2009iris, abdullah2016robust}, which iteratively evolve a curve under constraints in order to fit non-circular and non-elliptical boundries. These models, while more flexible than their predecessors, require careful initialization and can be computationally intensive. OSIRIS \cite{othman2016osiris} is an open-source software for iris recognition, using Hough transform and Gabor filters for accurate iris segmentation. It serves as a benchmark in evaluating new iris segmentation techniques. SegNet \cite{badrinarayanan2017segnet} is another well-known segmentation method that emerged as a notable deep learning architecture. Originally developed for scene understanding, SegNet’s efficient encoder-decoder structure makes it well-suited for iris
segmentation tasks. It effectively captures contextual information and detailed spatial resolution, crucial for accurately delineating the iris from other ocular components. The U-Net architecture, a variant of CNNs, has gained traction for its efficacy in segmenting medical images, including the iris \cite{lian2018attention}, by using a contracting path to capture context and a symmetrically expanding path that enables precise localization \cite{RFB15a}.

\section{Proposed Method} 

In this work, the SAM technique, renowned for its adaptability in medical image segmentation \cite{ma2024segment}, has been fine-tuned for the specific task of iris segmentation. In the context of iris segmentation, the bounding box prompt emerges as the most effective, providing clear demarcation of the Region of Interest (ROI) with minimal user input. This approach is not only intuitive but also aligns with standard ophthalmic measurement practices, ensuring both accuracy and efficiency \cite{ma2024segment}. In the context of iris segmentation using SAM, bounding boxes are crucial for directing the model's focus to the relevant region (the iris) in each image. These bounding boxes are derived from ground truth masks during training using a specialized function that calculates the minimum and maximum coordinates encompassing the iris, with random adjustments in perturbed bounding boxes that entail unpredictably altering their size and position. This randomness enhances a model's flexibility and generalization to new data. This approach ensures that the model consistently focuses on the iris region across different images. \emph {(Note that the proposed method does not require bounding boxes as input during the inference stage.)} 

In iris segmentation, one often encounters a significant class imbalance. That is, the number of pixels belonging to the target object (iris pixels) might be much smaller compared to the background (non-iris pixels). Traditional loss functions, such as Cross-Entropy or Mean Squared Error, when used in such scenarios, tend to get overwhelmed by the sheer number of negative samples (non-iris pixels), leading the model to produce trivial solutions that identify everything as the background (\Cref{fig:finetunes-casia}).

Focal Loss, introduced by Lin et al. \cite{8417976}, provides a refined approach to handle the prevalent issue of class imbalance. Rather than treating all misclassified samples uniformly, the Focal Loss assigns varying weights to samples based on their misclassification levels (gives more weight to samples that the model struggles to identify as either part of the iris or non-iris, and less weight to samples that the model finds easy to categorize as belonging to the iris or non-iris regions). This is done to focus the model's learning more on challenging cases. Focal Loss is derived from standard cross-entropy loss \cite{cover1999elements} that is inadequate to address this imbalance, predominantly focusing on the majority class and, thus, overlooking the minority class, i.e., the iris pixels in our case. 

Below, we see the cross-entropy (CE) loss for binary classification.
For a given probability $p$ of the positive class (iris) and the ground truth label $y$, the CE loss is defined as \cite{8417976}:

\begin{equation}
    CE(p,y) = 
    \begin{cases}
        -\log(p), & \text{if } y=1 \\
        -\log(1-p), & \text{otherwise.}
    \end{cases}
\end{equation}
For a pixel $p_{i}$, the cross-entropy loss is given by:
\begin{equation}
L_{CE}(p_i) = -y_i \log(\hat{y}_i) - (1 - y_i) \log(1 - \hat{y}_i),
\end{equation}

\noindent where, $y_i$ is the ground truth (1 for iris, 0 for background) and $\hat{y}_i$ is the predicted probability for the pixel being part of the iris. However, in such scenarios with a large class imbalance, the CE loss can be overwhelmed by a great number of easy-to-classify negative samples (non-iris pixels). As a result, we need to have balancing and focusing factors to solve the issue.

Focal Loss is introduced to down-weight the correctly classified pixels, thus forcing the model to focus on hard-to-classify pixels. To address the limitations of cross-entropy loss in the presence of class imbalance, Focal Loss, denoted as $FL(p_t)$, is introduced \cite{8417976}. It is formulated as:

\begin{equation}
    FL(p_t) = -(1 - p_t)^{\gamma} \log(p_t).
\end{equation}

$p_t$ is the model's estimated probability for the class with label $y=1$.
$\alpha$ is a weighting factor to balance the importance of positive vs. negative examples and $\gamma$ is the focusing parameter, adjusting the rate at which easy examples are down-weighted.
For our problem, we adapt it as follows: \\

If the ground truth label $y = 1$ (iris), then $p_t = \hat{y}_i$; otherwise $p_t = 1 - \hat{y}_i$. The modified Focal Loss for a pixel $p_i$ becomes: 
\begin{equation}
L_{\text{Focal}}(p_i) = -\alpha (1 - p_t)^\gamma y_i \log(p_t) - (1 - \alpha) p_t^\gamma (1 - y_i) \log(1 - p_t).
\end{equation}

Focal Loss, with its modulating factor $(1 - p_t)^\gamma$ diminishes the contribution of well-classified (easy) pixels and amplifies the loss for misclassified ones, especially near the iris boundary. Here, \(\alpha\) is a balancing factor, and \(\gamma\) is the focusing parameter that determines how much the weight of well-classified examples should be reduced. As \(\gamma\) increases, the effect of the modulating factor becomes more pronounced. The intuition behind Focal Loss can be visualized in the iris segmentation task. In an eye image, vast areas are not part of the iris (sclera, pupil, eyelids, etc.). When using standard cross-entropy, the model quickly learns to predict these easy, non-iris regions correctly, while struggling with the actual iris region due to class imbalance. The function comprises two parts:

\begin{itemize}[leftmargin=*]
    \item The first part, \( -\alpha(1 - p_t)^\gamma y_i\log(p_t) \), focuses on the positive class. The modulating factor \( (1 - p_t)^\gamma \) increases the loss for misclassified or uncertain positive examples.
    \item The second part, \( -(1 - \alpha) p_t^\gamma (1 - y_i) \log(1 - p_t) \), focuses on the negative class. The modulating factor \( p_t^\gamma \) increases the loss for misclassified or overconfident negative examples.
\end{itemize}

\begin{figure}[t]  
    \centering
    \includegraphics[width=0.93\linewidth]{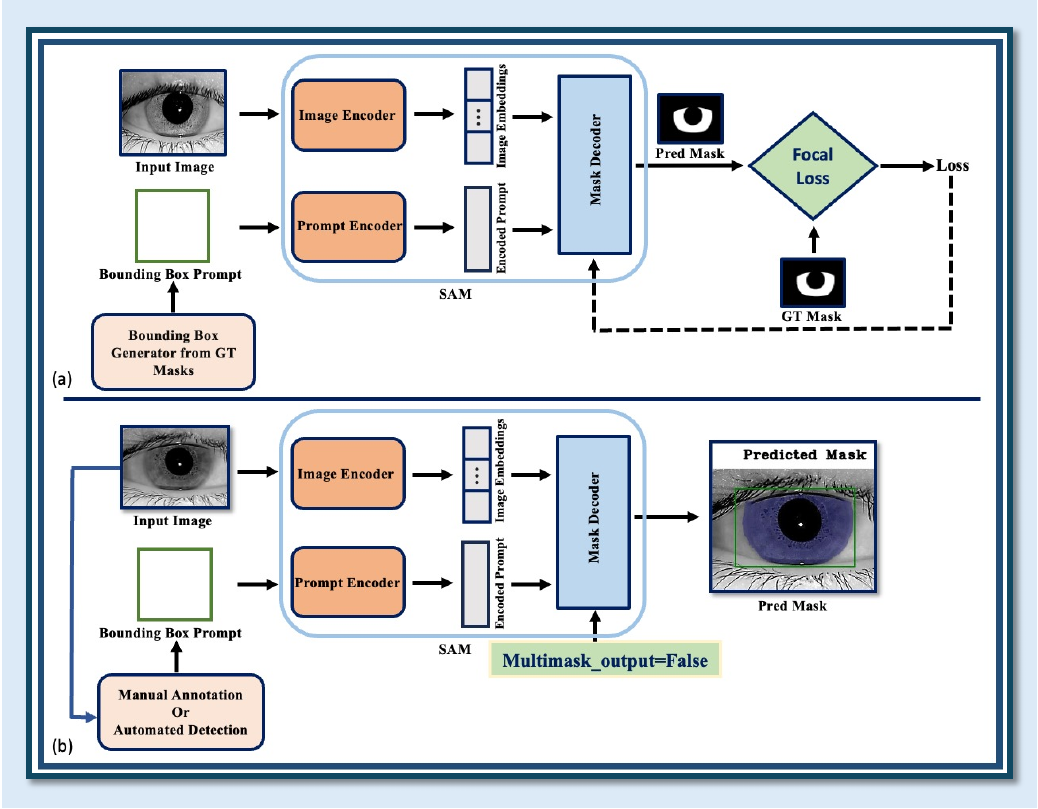} 
    \vspace{+2mm}
  \captionsetup{labelsep=period} 
  \captionsetup{labelfont=bf}    
  \caption{Proposed network (Iris-SAM) using Segment Anything Model (SAM) \cite{kirillov2023segment}. (a) Training and (b) Inference/Testing. During training, prompts (bounding boxes) are generated from ground truth masks to guide the model. For inference/testing, the model automatically generates bounding boxes (visualized in green) from the input image, allowing it to predict the iris masks (depicted in blue) without needing explicit bounding box inputs.}
    \label{fig:architecture}
\end{figure}

By adjusting \( \alpha \) and \( \gamma \), Focal Loss can be fine-tuned for different datasets and imbalance problems, aiming to improve the performance on hard examples. During the training phase, we utilize a dataset consisting of iris images $I_1, I_2, \ldots, I_n$, each paired with a specific bounding box $B_1, B_2, \ldots, B_n$ and a corresponding ground truth mask $M_1, M_2, \ldots, M_n$. The model, when presented with an image $I_j$, produces a predicted mask $O_j$. A key objective is to enhance the agreement between $O_j$ and the ground truth mask $M_j$, with the Intersection over Union (IoU) metric serving as the measure of this congruence. The IoU is defined as $\text{IoU}(O_j, M_j) = \frac{|O_j \cap M_j|}{|O_j \cup M_j|}$, indicating the extent of overlap between the predicted and actual iris regions (the standard deviation of IoU across all images is also computed, offering insight into the consistency of the model’s performance). Crucially, the difference between the predicted mask $O_j$ and the ground truth mask $M_j$ plays a pivotal role in the training process. This disparity indicates how much the model's predictions deviate from the real iris area, forming a basis for calculating the weights used in the Focal Loss function. These weights, derived from the degree of misalignment between $O_j$ and $M_j$, are employed to adjust the model’s learning focus, particularly directing attention toward more challenging segmentation areas. In Focal Loss, the weights for each example are dynamically calculated using the formula $\alpha \times (1 - p_t)^\gamma$, where $p_t$ is derived from the model's output. The model's output logits are first transformed into a probability $p$ using the sigmoid function. Then, $p_t$ is determined based on the ground truth label $y$: if $y = 1$, $p_t = p$; if $y = 0$, $p_t = 1 - p$. This approach ensures that the loss focuses more on challenging examples where the model is less confident. This adjustment is vital for refining the accuracy of the model's mask decoder, as it guides the model to learn more effectively from areas where its predictions are less accurate, thereby enhancing its overall segmentation capability.



SAM's versatile architecture allows for the generation of multiple masks corresponding to various objects or regions within an image. However, to tailor its capability to our specific need of segmenting a single object (the iris), we utilize the \texttt{multimask\_output} parameter within the model's prediction method. The parameter is a boolean flag that guides the model's output behavior.
As can be seen in \Cref{fig:architecture}, setting \texttt{multimask\_output} to \texttt{False} instructs SAM to deviate from its default behavior of identifying multiple objects. Instead, it concentrates on generating a singular focused mask output. This output, in the context of our work, is the segmented iris region.
This focused approach of generating a single mask output ensures that the model's computational resources and learning capacity are entirely devoted to the task of iris segmentation. It negates the diversion of attention to multiple regions, thereby enhancing the precision and accuracy of the segmentation process. This modification is pivotal for the specific requirements of iris biometric systems, where the delineation precision of the iris region directly impacts the system's reliability and performance.

The data split information for the experiments can be seen in \Cref{tab:dataset_info}. Datasets are captured using diverse camera systems and under varying environmental conditions, resulting in a wide range of image qualities, iris textures, and lighting variations. In our experiments, the SGD optimizer with a learning rate of 0.0001 and training for 100 epochs was optimal, achieving convergence without overfitting. This configuration not only optimized the learning curve but also stabilized the loss across epochs, as seen in the accompanying plots (\Cref{fig:two-col-2}).


\section{Experimental Results}
\subsection{Dice, Triplet, and Focal Losses}
\begin{figure}[t]
  \centering
  \begin{subfigure}[b]{0.49\linewidth}
    \includegraphics[width=\linewidth]{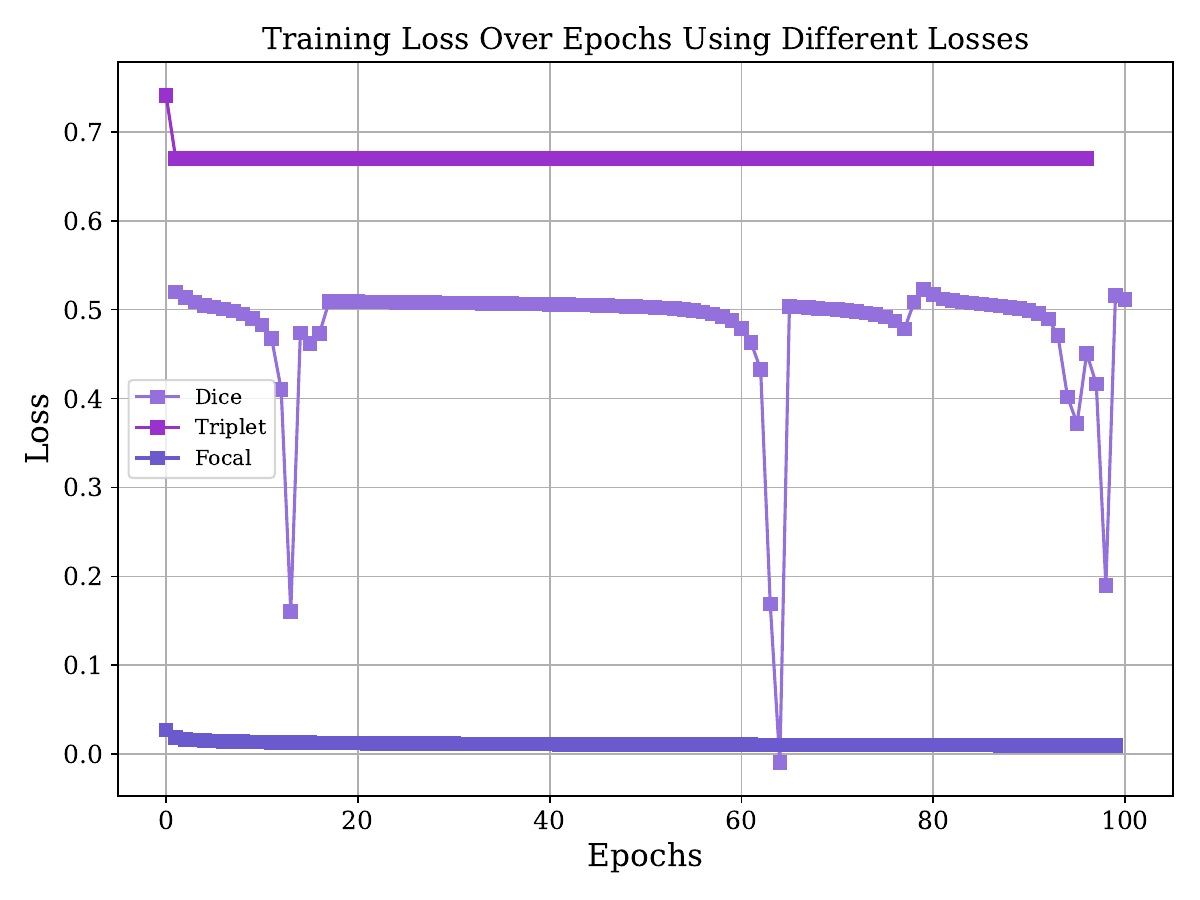}
    \caption{} 
    \label{fig:3loss}
  \end{subfigure}
  \hfill 
  \begin{subfigure}[b]{0.49\linewidth}
    \includegraphics[width=\linewidth]{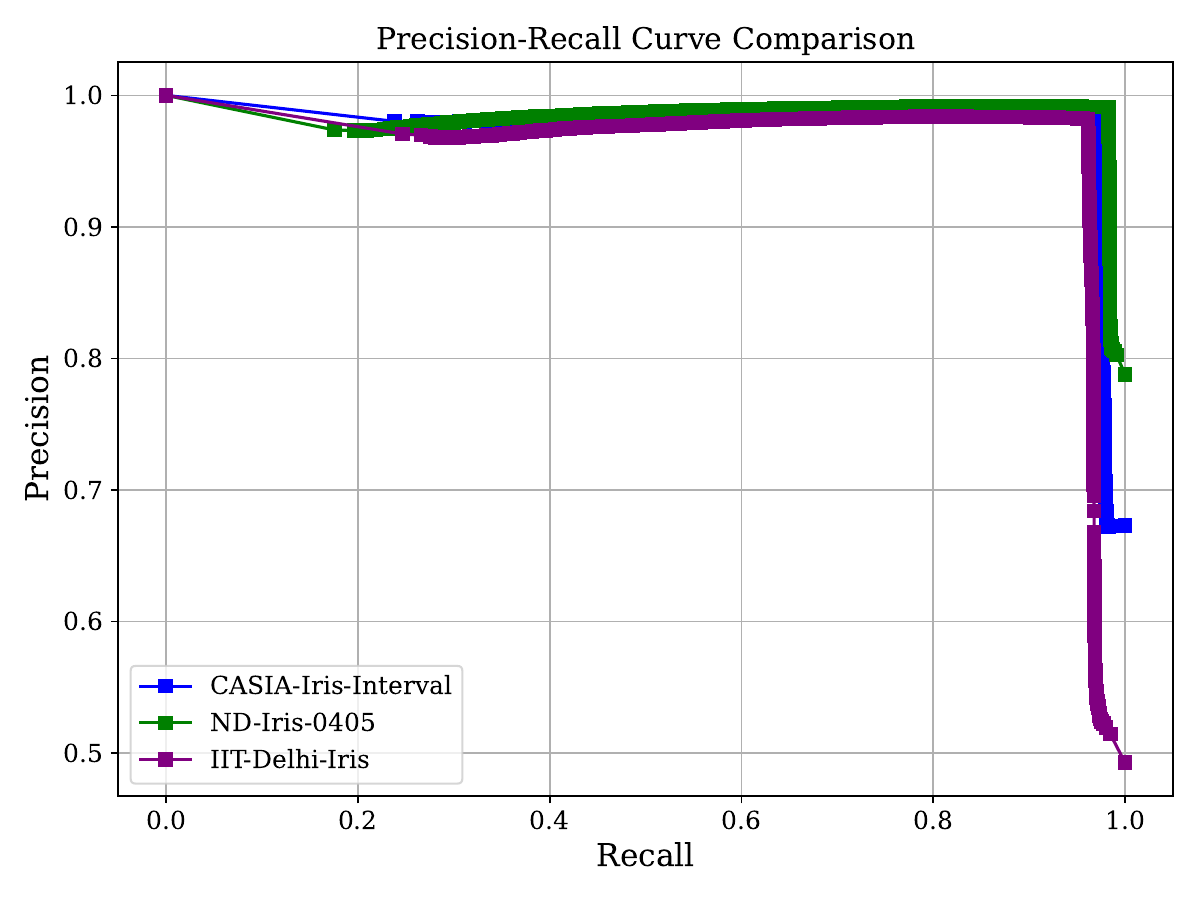}
    \caption{} 
    \label{fig:recall}
  \end{subfigure}
  \captionsetup{labelsep=period} 
  \captionsetup{labelfont=bf}
  \caption{(a) Training loss over 100 epochs for Dice, Triplet, and Focal Loss functions on the Casia-Iris-Interval-v3 dataset. Focal Loss converges rapidly to a lower value, indicating efficient learning, while Dice and Triplet losses exhibit higher variability and slower convergence. (b) Precision-Recall curve of our method on different test datasets when the Focal Loss was used during training.}
  \label{fig:two-col-1}
\end{figure}

In our iris segmentation experiments, Focal Loss outperformed Dice and Triplet losses, excelling in managing class imbalance by focusing on difficult examples, as shown in our training loss graph (\Cref{fig:3loss}). This approach effectively emphasized critical iris details and boundaries, leading to more accurate segmentation. Focal Loss achieved lower, stable loss values rapidly, indicating efficient learning. Conversely, Dice Loss showed higher, fluctuating loss values, struggling with false positives and false negatives, while Triplet Loss, proved ineffective for segmentation, lacking in pixel-wise accuracy. 
\begin{table}[t]
\centering
\captionsetup{labelsep=period} 
\captionsetup{labelfont=bf}
\caption{Dataset information.}
\label{tab:dataset_info}
\resizebox{\textwidth}{!}{
\begin{tabular}{lccc}
\toprule
\textbf{Dataset} & \textbf{Total} & \textbf{Train} & \textbf{Test} \\
 & \textbf{(Images/Identities)} & \textbf{(Images/Identities)} & \textbf{(Images/Identities)} \\
\midrule
CASIA-Iris-Interval-v3~\cite{casiairisv3} & 2655/249 & 2294/199 & 361/50 \\
ND-Iris-0405~\cite{bowyer2016nd}      & 804/346  & 624/276  & 180/70 \\
IIT-Delhi-Iris~\cite{kumar2010comparison}     & 2240/224 & 1790/179 & 450/45 \\
\bottomrule
\end{tabular}
} 
\end{table}

\subsection{Segmentation Results}
\begin{figure}[t]
  \centering
  \begin{subfigure}[b]{0.49\linewidth}
    \includegraphics[width=\linewidth]{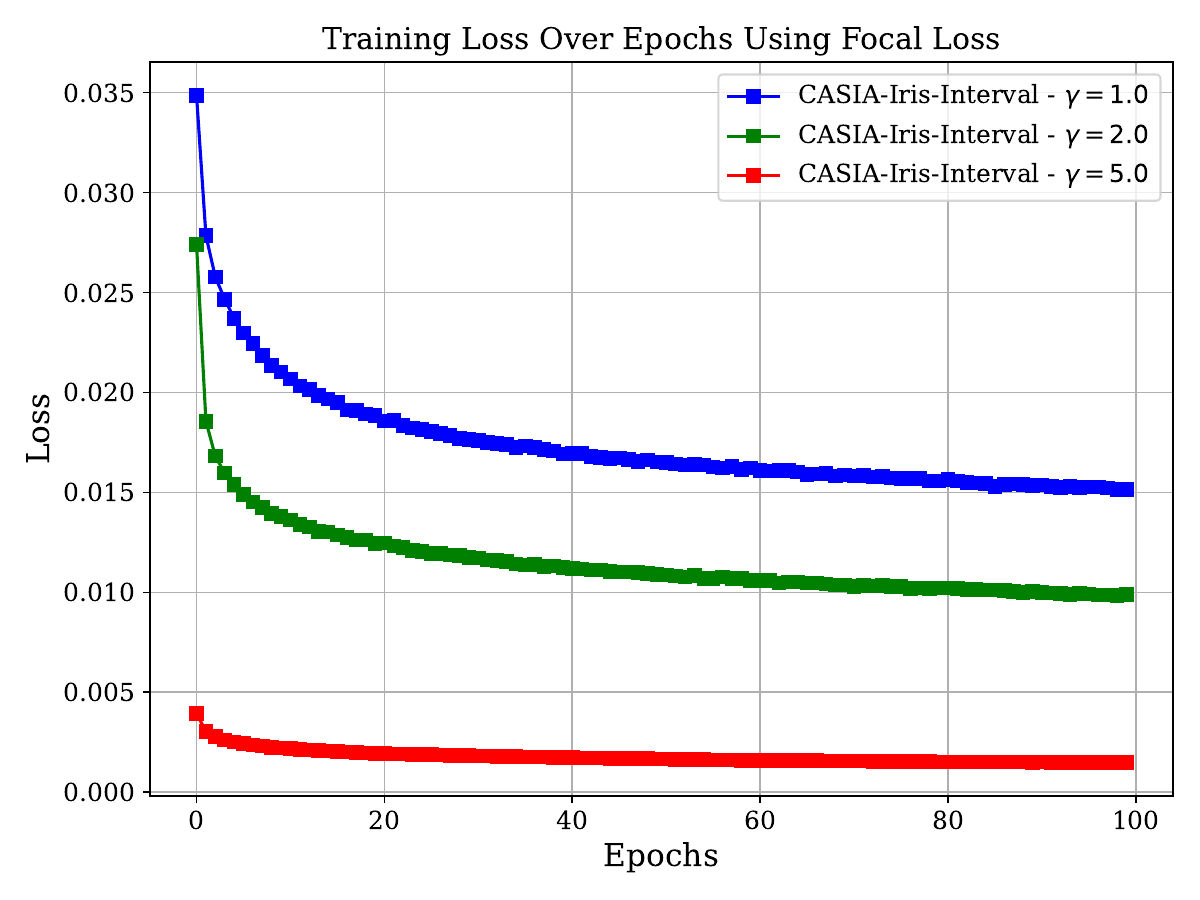}
    \caption{} 
    \label{fig:fig7a}
  \end{subfigure}
  \hfill 
  \begin{subfigure}[b]{0.49\linewidth}
    \includegraphics[width=\linewidth]{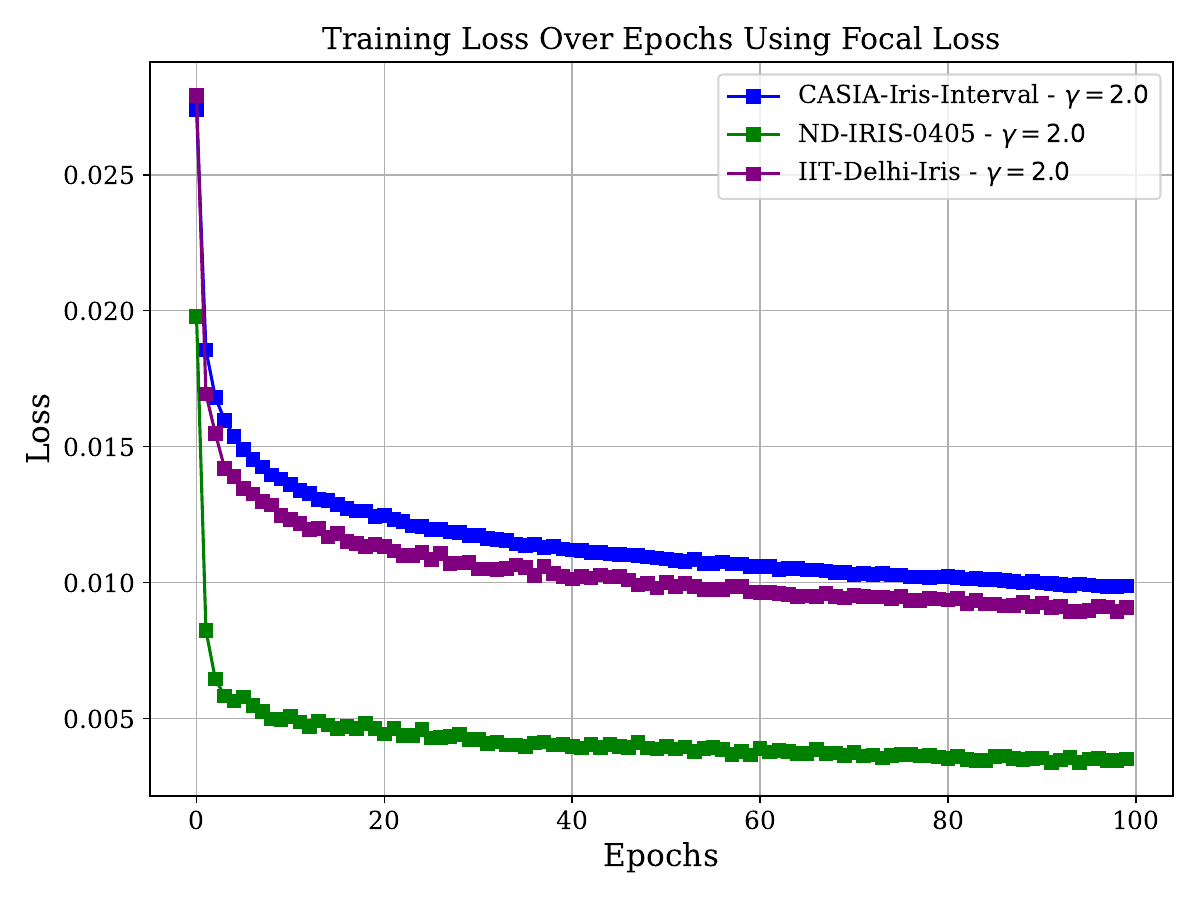}
    \caption{} 
    \label{fig:nd-casia}
  \end{subfigure}
  \captionsetup{labelsep=period} 
  \captionsetup{labelfont=bf}
  \caption{(a) FineTuning + Focal Loss (Iris-SAM) using the default pre-trained model "ViT\_h" with different $\gamma$ values on CASIA-Iris-Interval-v3 dataset. (b) FineTuning + Focal Loss (Iris-SAM) using the default pre-trained model "ViT\_h" on three different datasets.}
  \label{fig:two-col-2}
\end{figure}

Our experimentation spanned across various settings of the focusing parameter, $\gamma$, within the Focal Loss, examining its impact on the model's performance. We tested $\gamma$ values of 1, 2, and 5 on the CASIA-Iris-Interval-v3 dataset (\Cref{fig:fig7a}), aiming to calibrate the loss function's sensitivity to misclassified instances. A $\gamma$ value of 2 emerged as the optimal choice, striking a delicate balance between learning efficiency and robustness (\Cref{fig:nd-casia}). While a higher $\gamma$ of 5 initially seemed promising, it ultimately led to overfitting--a testament to the nuanced trade-offs in model training. The overfitting was evidenced by a good performance on the training set that did not generalize well to unseen data. 

As can be seen in \Cref{tab:average_iou}, a $\gamma$ value of 2, achieved a high Average IoU of 96.94\% on the CASIA-Iris-Interval-v3 and 99.58\% on the ND-Iris-0405 dataset, demonstrating its effectiveness in iris segmentation. However, performance slightly dipped on the IIT-Delhi-Iris dataset with an Average IoU of 94.34\%, indicating a nuanced variance in efficacy across different datasets. This marginal drop in performance can be attributed to the dataset's unique characteristics, particularly, the prevalence of dense eyelashes. The eyelashes present a complex segmentation challenge. Our model's nuanced attempt to delineate these eyelashes more accurately contributed to the slight reduction in the overall IoU score. It is a reflection of the model's sensitivity to fine-grained features, a trait that is both a strength and a point of careful consideration in dataset diversity.

The good performance of our model is further underscored by the standard deviation associated with the Average IoU scores, which is between 0.002 to 0.008 across the evaluated datasets. The limited variance in the IoU scores suggests a high level of consistency and reliability in the model's ability to segment iris regions, despite the inherent challenges presented by different images. Such consistency is particularly noteworthy when considering the complexity of iris textures and the potential for occlusions like eyelashes and reflections, which typically introduce variability in segmentation tasks.

\begin{figure}[H]
    \centering
    \begin{subfigure}[b]{0.32\textwidth}
        \includegraphics[width=\textwidth]{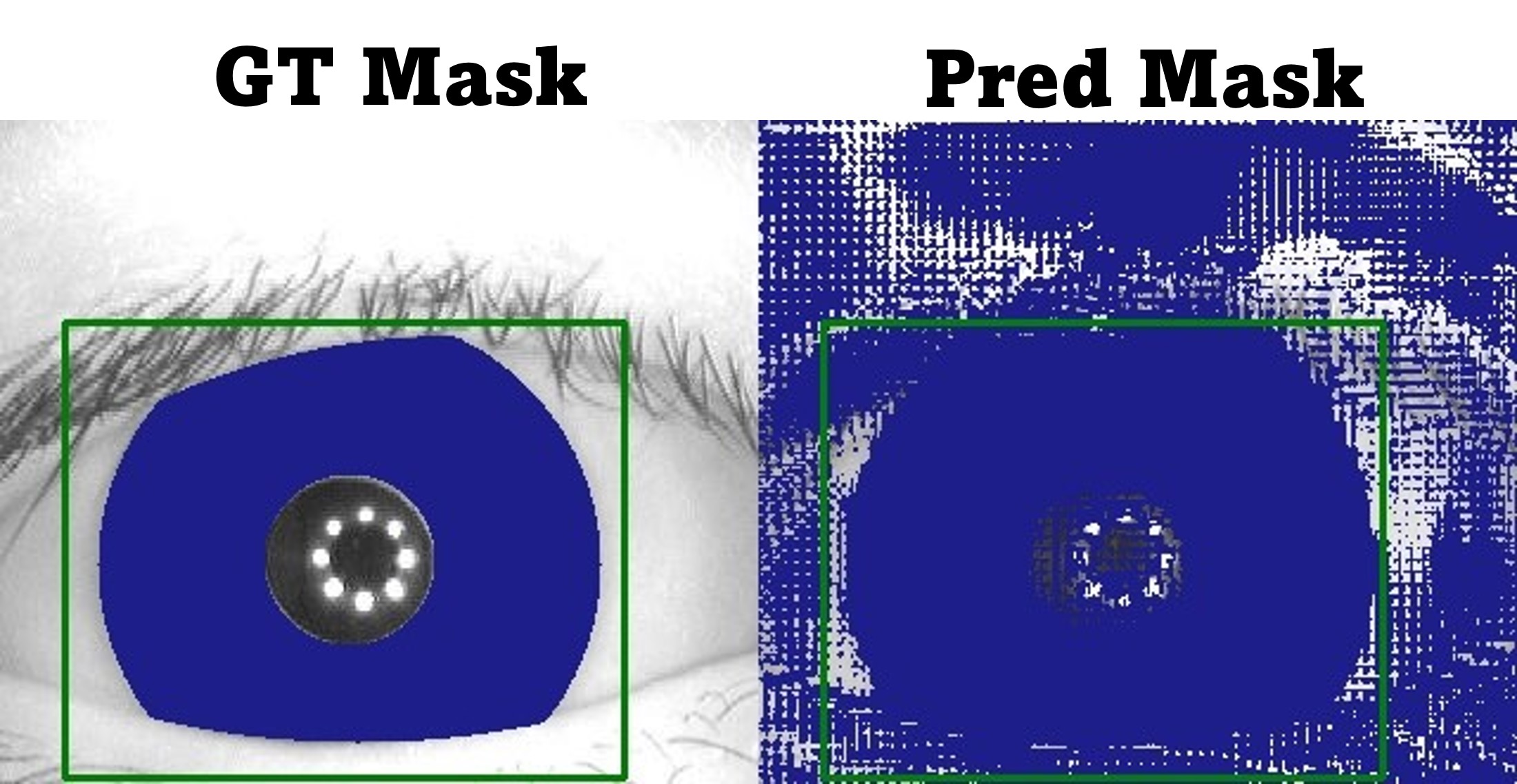}
        \caption{FT Only - S1002R10}
        \label{fig:finetune1-S1002R10}
    \end{subfigure}
    \begin{subfigure}[b]{0.32\textwidth}
        \includegraphics[width=\textwidth]{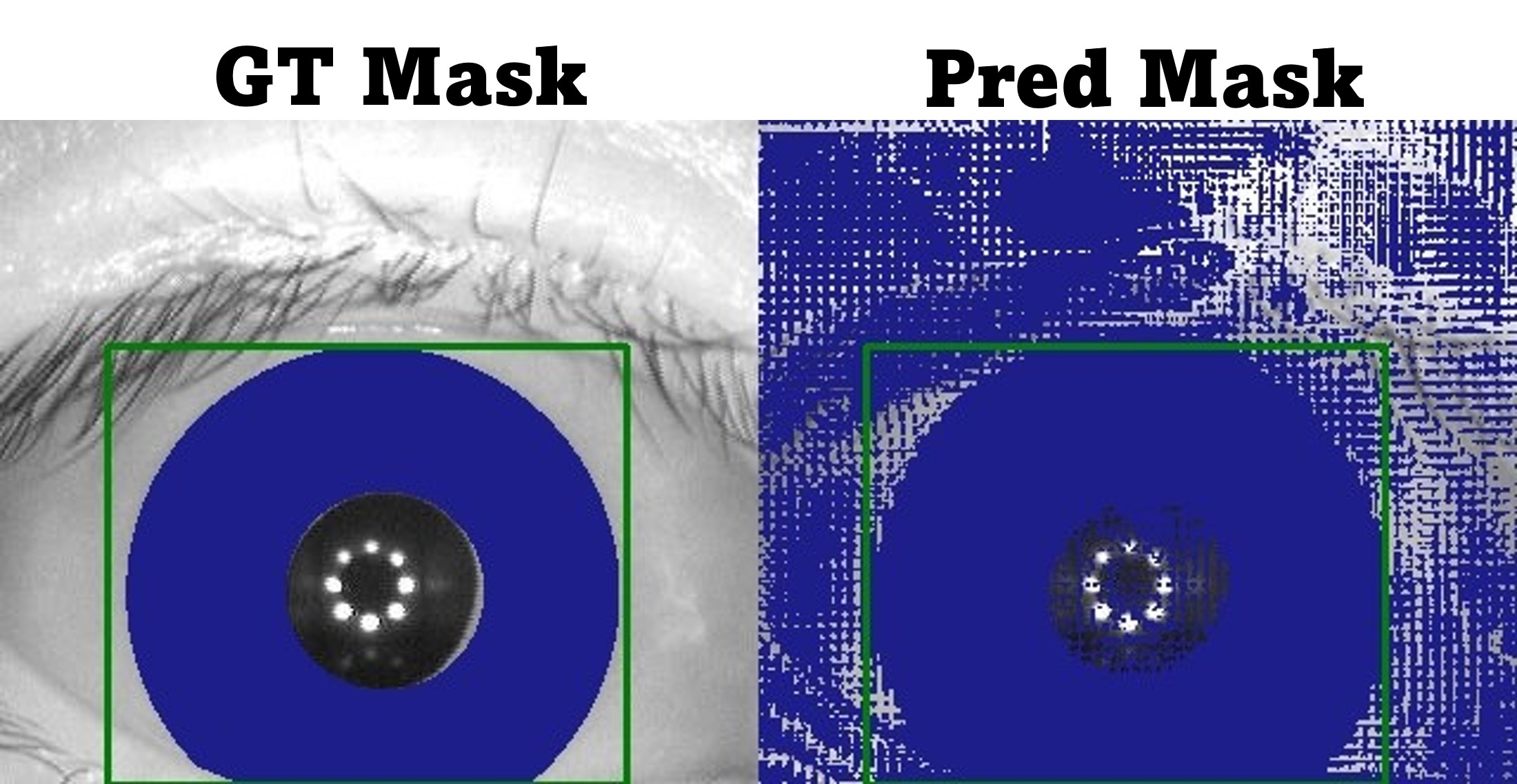}
        \caption{FT Only - S1019R04}
        \label{fig:finetune2-S1019R04}
    \end{subfigure}
    \begin{subfigure}[b]{0.32\textwidth}
        \includegraphics[width=\textwidth]{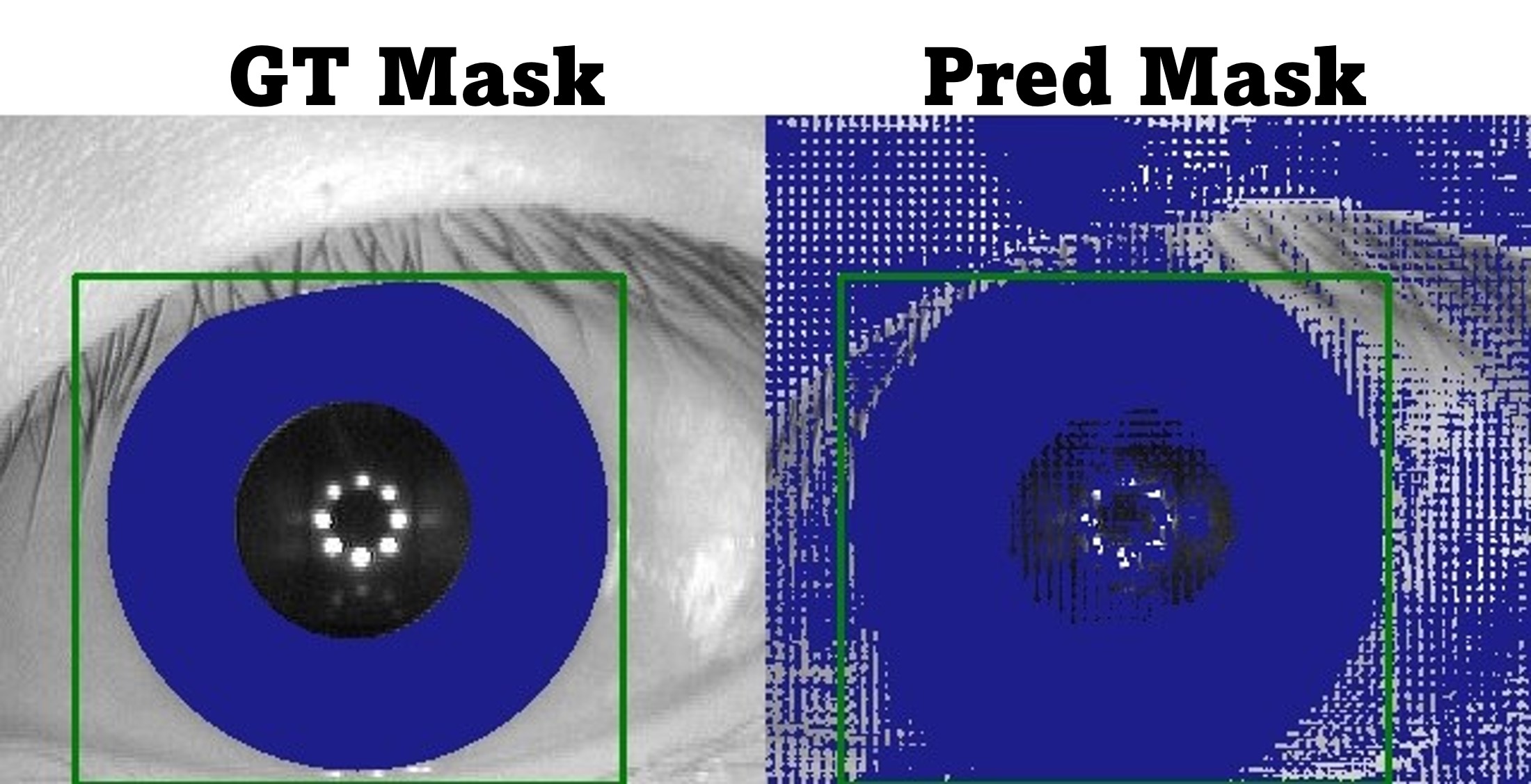}
        \caption{FT Only - S1053L05}
        \label{fig:finetune3-S1053L05}
    \end{subfigure}
    \captionsetup{labelsep=period} 
  \captionsetup{labelfont=bf}
    \caption{FineTuning (FT) sample results (CASIA-Iris-Interval-v3).}
    \label{fig:finetunes-casia}
\end{figure}

\begin{figure}[ht]
    \centering
    \begin{subfigure}[b]{0.32\textwidth}
        \includegraphics[width=\textwidth]{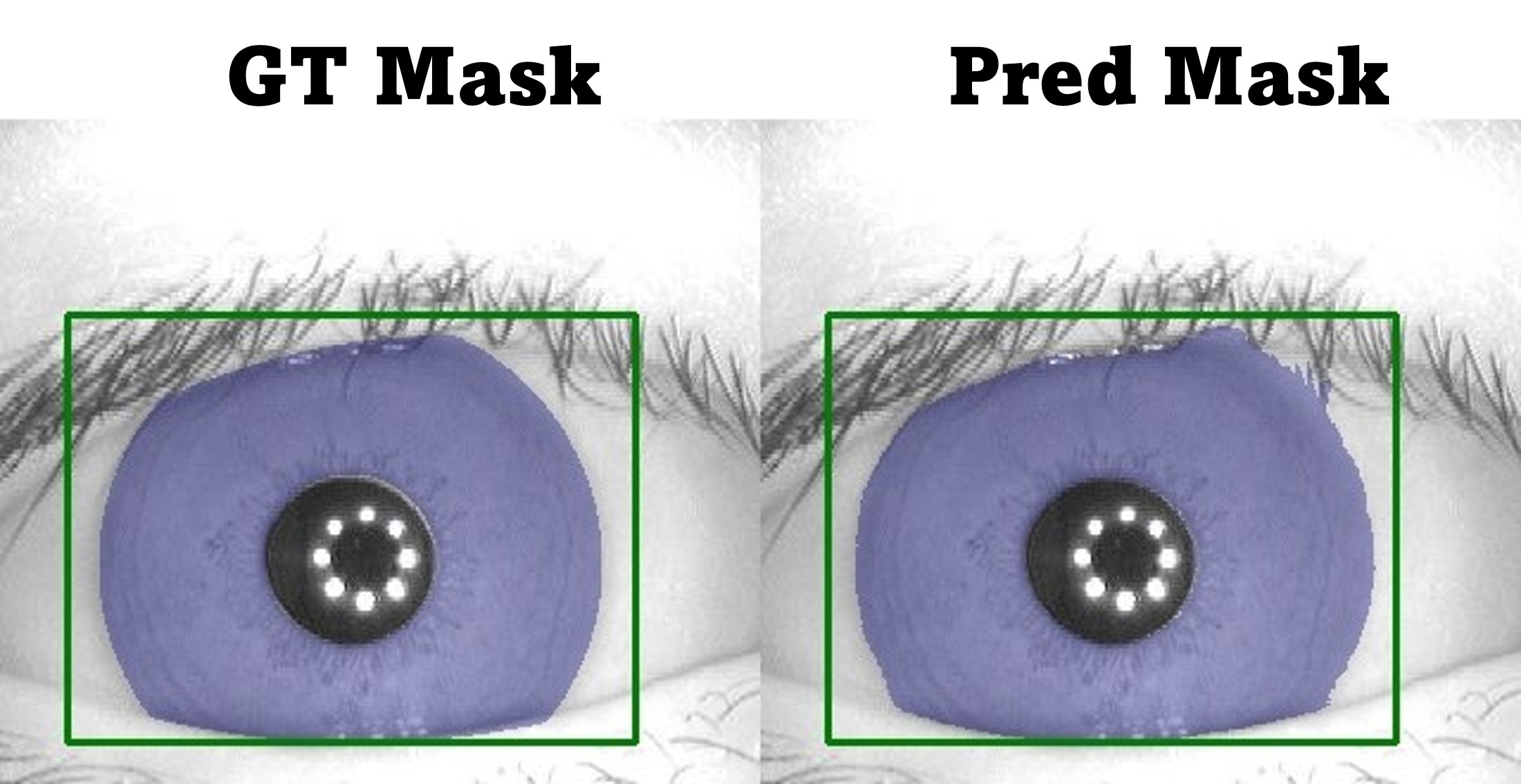}
        \caption{FT+Focal - S1002R10}
        \label{fig:focal1-S1002R10}
    \end{subfigure}
    \begin{subfigure}[b]{0.32\textwidth}
        \includegraphics[width=\textwidth]{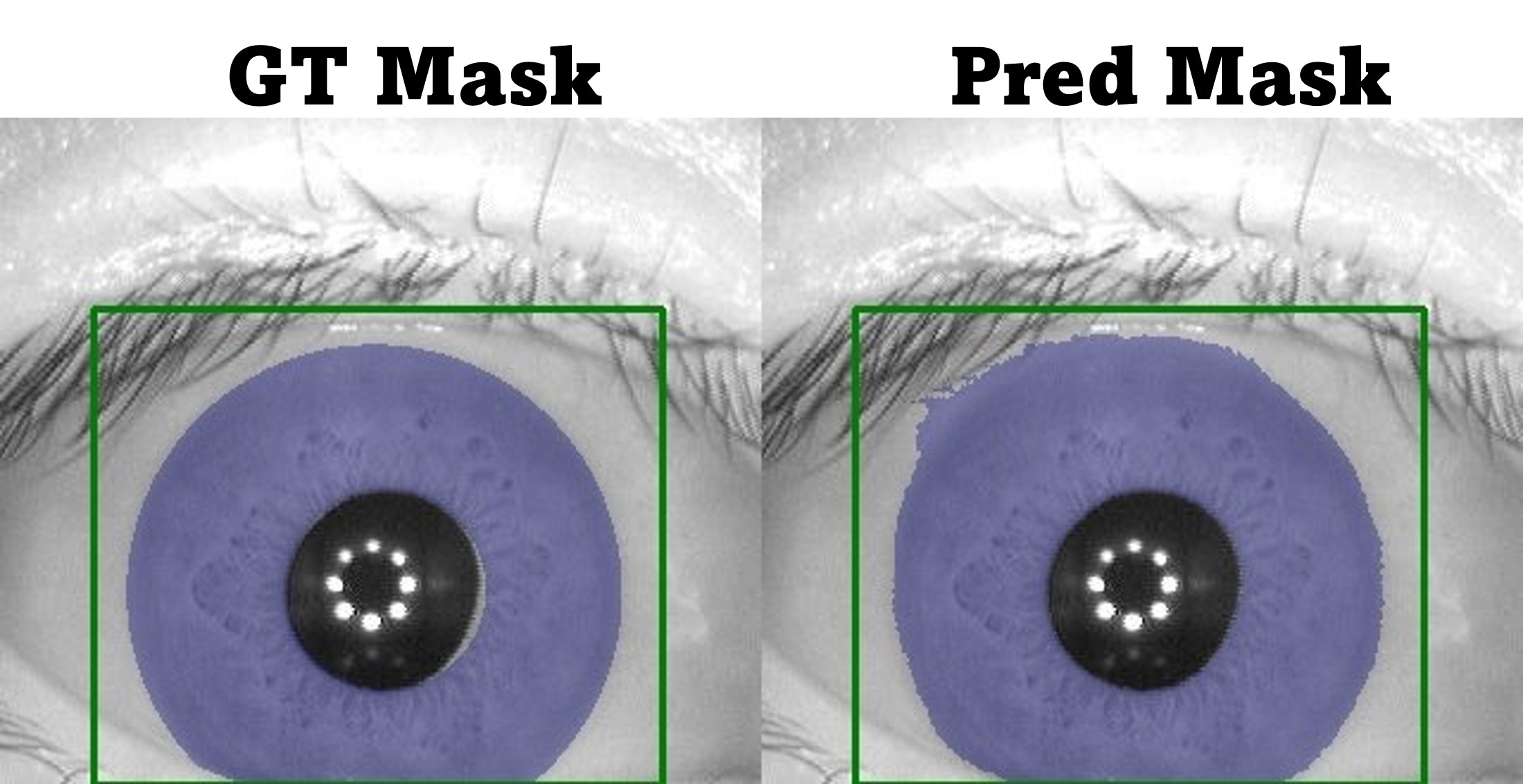}
        \caption{FT+Focal - S1019R04}
        \label{fig:focal2-S1019R04}
    \end{subfigure}
    \begin{subfigure}[b]{0.32\textwidth}
        \includegraphics[width=\textwidth]{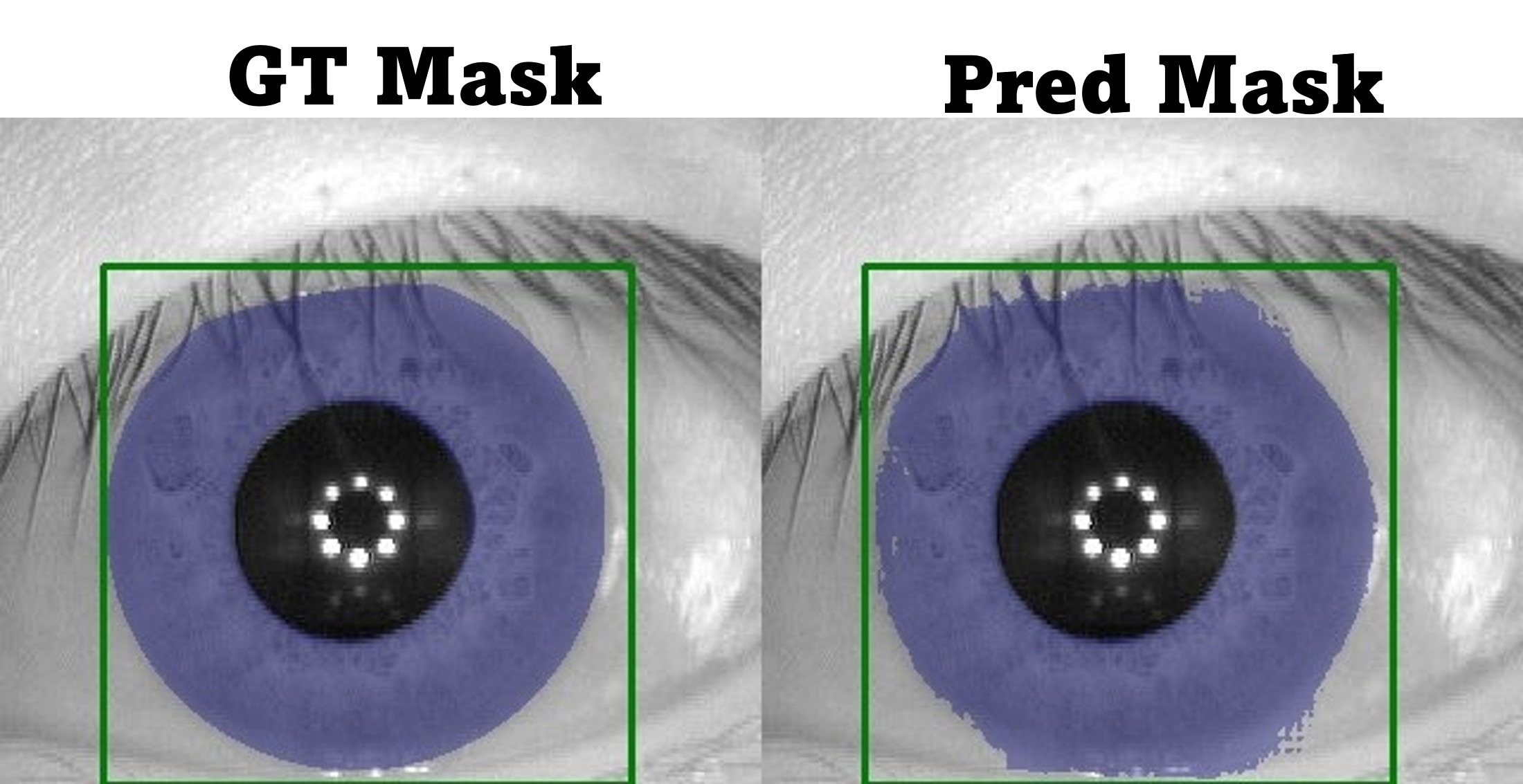}
        \caption{FT+Focal - S1053L05}
        \label{fig:focal3-S1053L05}
    \end{subfigure}
    \captionsetup{labelsep=period} 
  \captionsetup{labelfont=bf}
    \caption{FT + Focal Loss (Iris-SAM) sample results (CASIA-Iris-Interval-v3).}
    \label{fig:finetunes-focal-casia}
\end{figure}

\begin{figure}[H]
    \centering
    \begin{subfigure}[b]{0.32\textwidth}
        \includegraphics[width=1.0\textwidth]{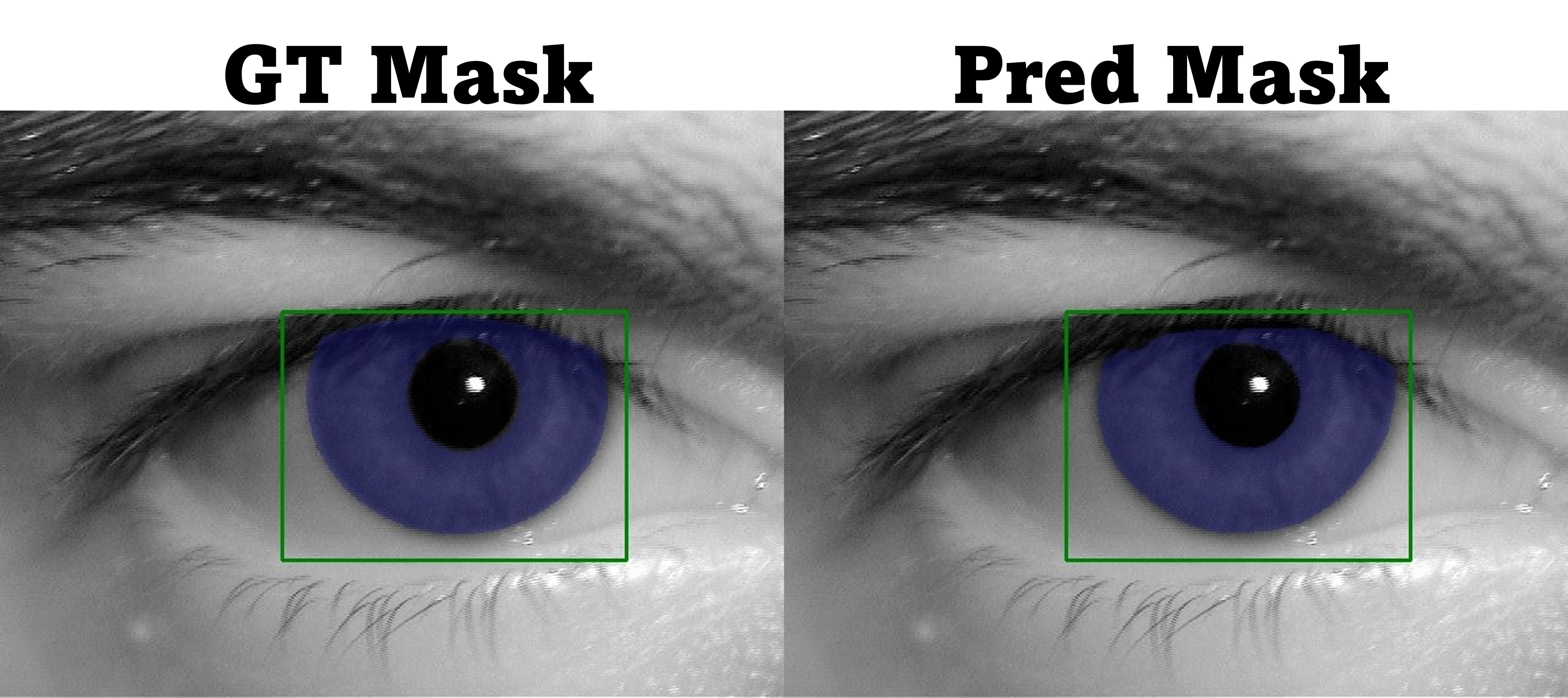}
        \caption{FT+Focal - 04267d227}
        \label{fig:focal1-04267d227}
    \end{subfigure}
    \begin{subfigure}[b]{0.32\textwidth}
        \includegraphics[width=1.0\textwidth]{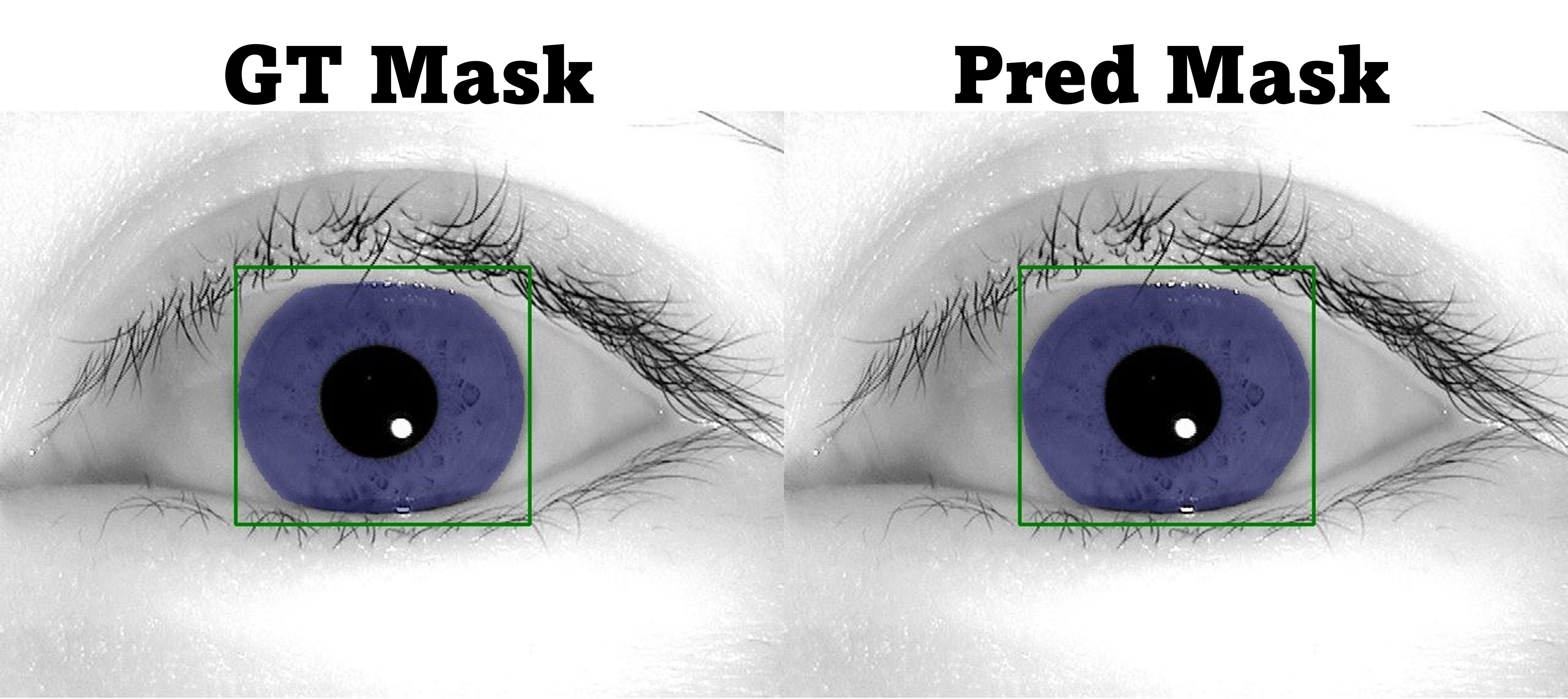}
        \caption{FT+Focal - 04350d458}
        \label{fig:focal2-04350d458}
    \end{subfigure}
    \begin{subfigure}[b]{0.32\textwidth}
        \includegraphics[width=1.0\textwidth]{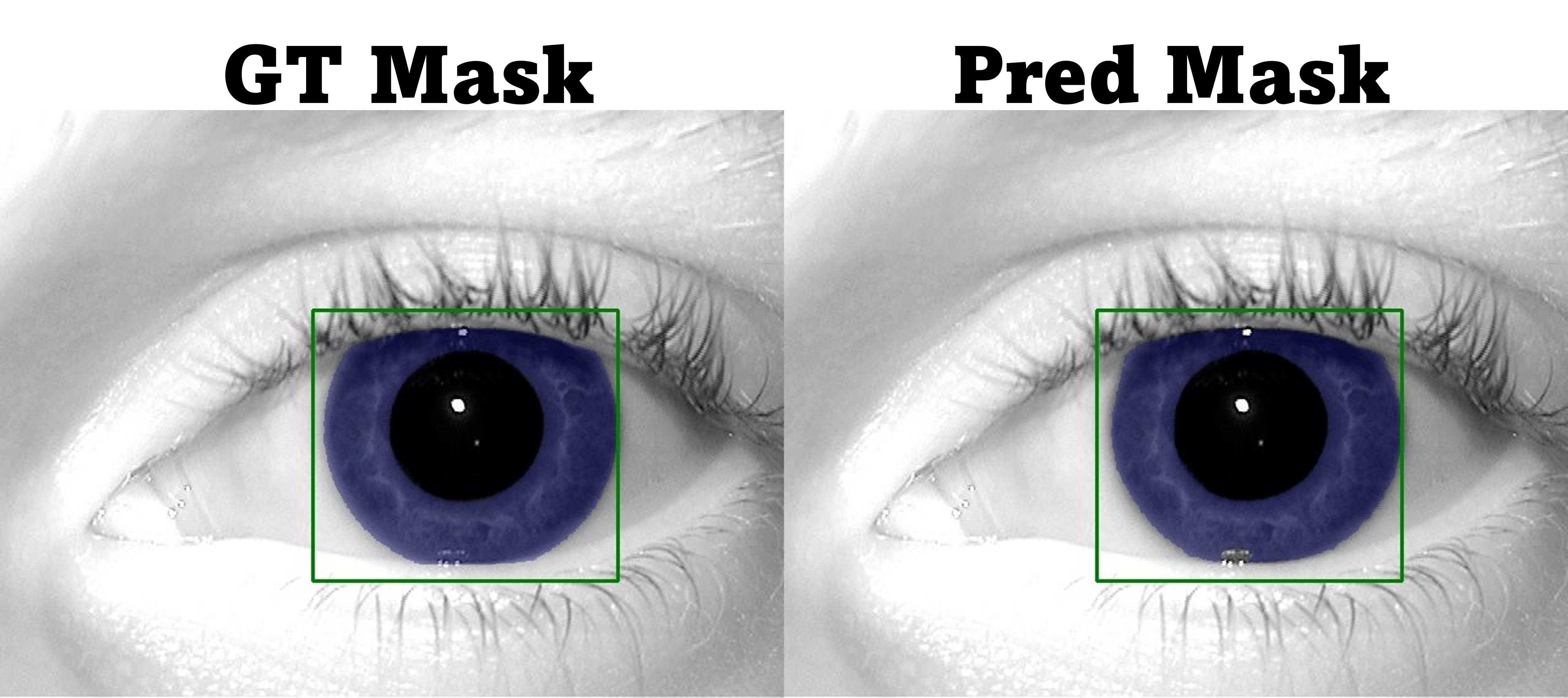}
        \caption{FT+Focal - 04370d361}
        \label{fig:focal3-04370d361}
    \end{subfigure}
    \captionsetup{labelsep=period} 
   \captionsetup{labelfont=bf}
    \caption{FT + Focal Loss (Iris-SAM) sample results (ND-IRIS-0405).}
    \label{fig:finetunes-focal-ndiris}
\end{figure}

\begin{figure}[H]
    \centering
    \begin{subfigure}[b]{0.32\textwidth}
        \includegraphics[width=1.0\textwidth]{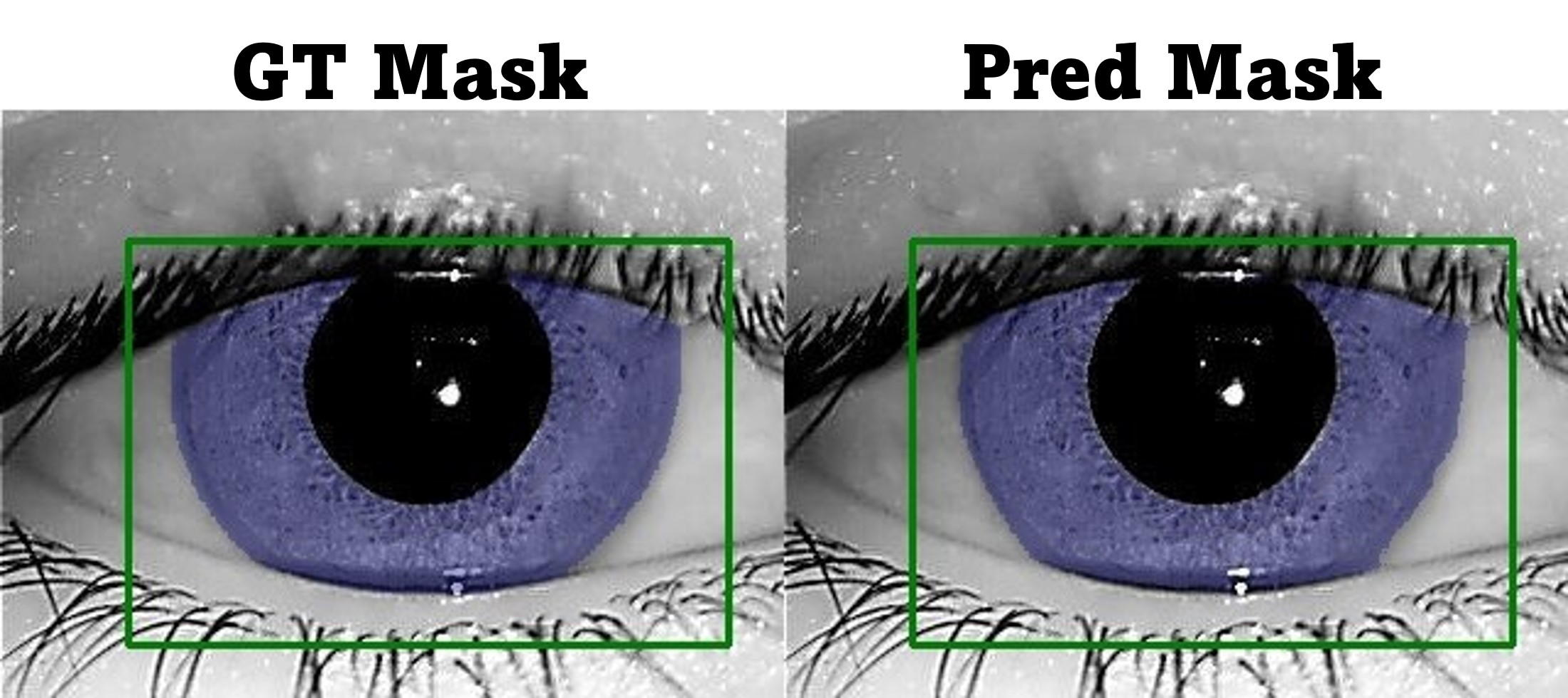}
        \caption{FT+Focal - 008\_07}
        \label{fig:focal1-008_07}
    \end{subfigure}
    \begin{subfigure}[b]{0.32\textwidth}
        \includegraphics[width=1.0\textwidth]{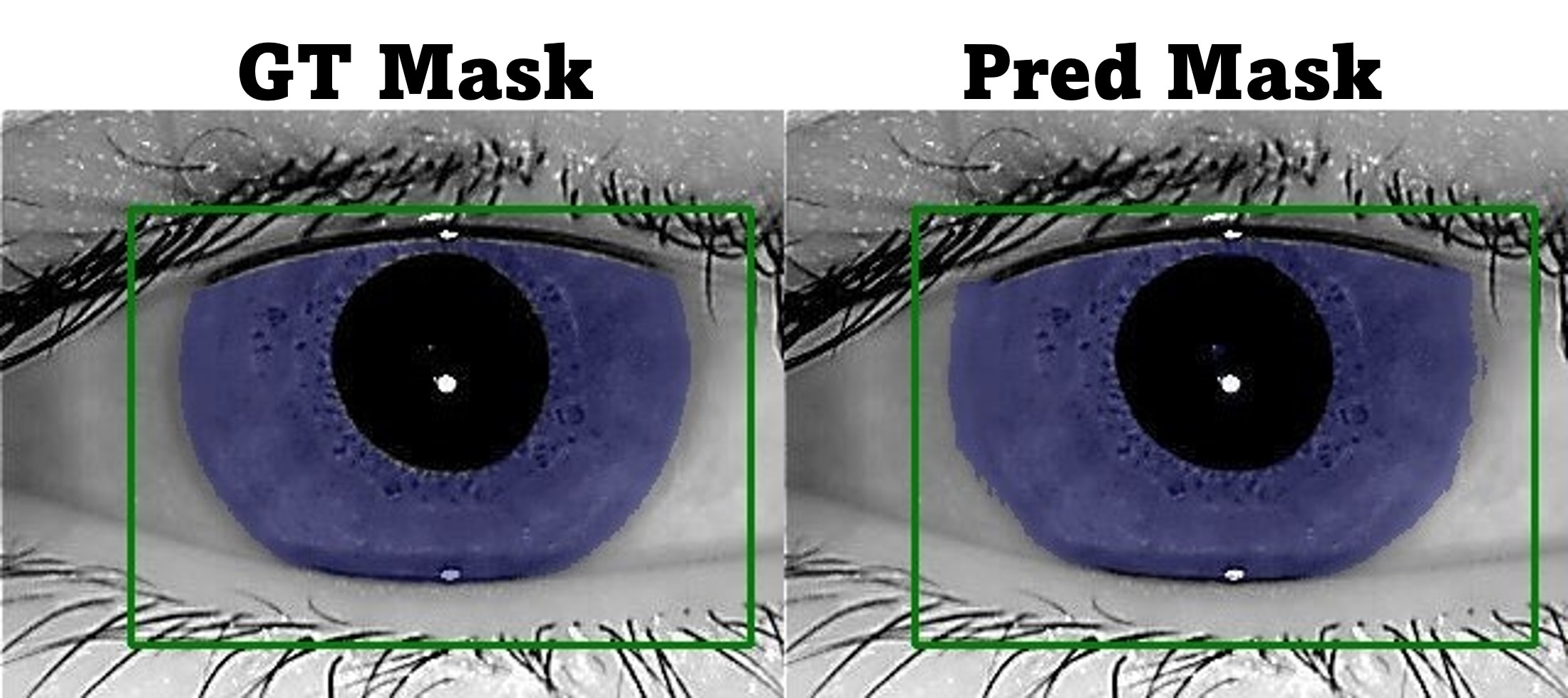}
        \caption{FT+Focal - 015\_04}
        \label{fig:focal2-015_04}
    \end{subfigure}
    \begin{subfigure}[b]{0.32\textwidth}
        \includegraphics[width=1.0\textwidth]{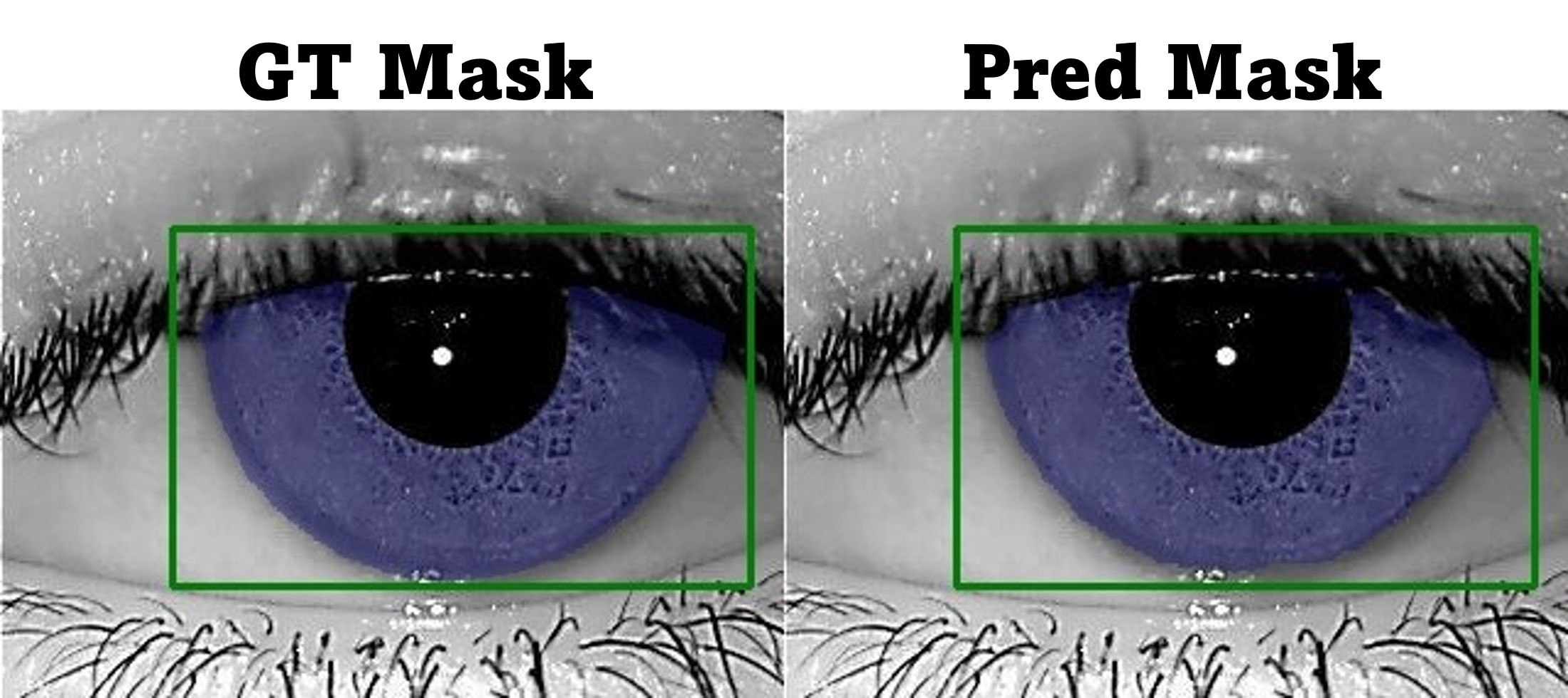}
        \caption{FT+Focal - 075\_06}
        \label{fig:focal3-075_06}
    \end{subfigure}
    \captionsetup{labelsep=period} 
    \captionsetup{labelfont=bf}
    \caption{FT + Focal Loss (Iris-SAM) sample results (IIT-Delhi-Iris).}
    \label{fig:finetunes-focal-iitd}
\end{figure}

The baseline section of \Cref{tab:average_iou} reports the performance of established iris segmentation models that are used as benchmarks. The OSIRIS model provides a baseline with an average IoU of 86.28\% (±6.50), while the DRN and Context-100k models show improved performance and consistency, with IoUs around 89\% and lower standard deviations. SegNet achieves an IoU of 89.75\% but with slightly higher variability. The numerical performance of the baseline methods further conveys the efficacy of the proposed method which has significantly higher IoU scores and near-zero standard deviation underscoring its advanced segmentation accuracy and consistency across various datasets. In \Cref{fig:finetunes-focal-casia,fig:finetunes-focal-ndiris,fig:finetunes-focal-iitd} we demonstrate the effectiveness of incorporating Focal Loss into our fine-tuning process for the Segment Anything Model (SAM). The bounding box is denoted in green and the pixels classified as iris are marked in blue. Prior to the integration of Focal Loss, the SAM's performance, while adequate, fell short of achieving the high levels of accuracy needed for precise iris segmentation (\Cref{fig:finetunes-casia}). The initial fine-tuning phase was unable to consistently differentiate the intricate iris patterns from surrounding ocular structures. However, the adaptation of Focal Loss marked a significant turning point in our experiments and allowed the SAM to focus more on discerning the nuanced boundaries of the iris. The post-Focal Loss fine-tuning results, as showcased in the figures, reveal a substantial alignment with the ground-truth data. The contrast between the pre and post-Focal Loss output is a testament to the loss function's impact. Where the original fine-tuning failed to capture certain iris details, the model augmented with Focal Loss succeeded in delineating the iris with remarkable precision. These results not only validate the rationale behind using the Focal Loss but also underscore its potential in handling diversity of pixels in both the positive and negative classes. 
\vspace{-1mm}

As depicted in \Cref{fig:bad-ground-truth1} and \Cref{fig:bad-ground-truth2}, our fine-tuned SAM model, with Focal Loss, exhibited the ability to not only match but sometimes surpass the ground-truth masks in accuracy. This was particularly evident when the model adeptly filled in missing iris segments and refrained from misclassifying eyelashes as iris tissue. To further validate our results, we plotted precision and recall curves for the three datasets (\Cref{fig:recall}), which reveal the model's strong segmentation capability, with ND-Iris-0405 showing near-perfect precision. Precision reflects the proportion of true positive pixels among the predicted positives, and recall represents the proportion of true positive pixels among the actual positives. While CASIA-Iris-Interval-v3 and IIT-Delhi-Iris experience minor precision declines at full recall, the consistently high performance indicates effective feature capture for iris segmentation.
\begin{table}[t]
\centering
  \captionsetup{labelsep=period} 
  \captionsetup{labelfont=bf}
\captionsetup{width=0.90\textwidth}
\caption{Iris segmentation accuracy in terms of Average IoU\% $\pm$ STD. Our method is observed to perform much better than the four baseline techniques. Further, it exhibits good generalization capability.}
\label{tab:average_iou}
\setlength{\tabcolsep}{12pt}
\begin{tabular}{|c|c|c|} 
\hline
\textbf{Our Method} & \textbf{Dataset} & \textbf{Accuracy} \\
\hline
Iris-SAM & CASIA-Iris-Interval-v3 & \textbf{96.94 $\pm$ 0.005} \\
Iris-SAM  & ND-Iris-0405 & \textbf{99.58 $\pm$ 0.003} \\
Iris-SAM & IIT-Delhi-Iris & \textbf{94.34 $\pm$ 0.008} \\
\hline
\multicolumn{3}{|c|}{\textbf{Our Method’s Generalization}} \\
\hline
\textbf{Train} & \textbf{Test} & \textbf{Accuracy} \\
\hline
ND-Iris-0405 & IIT-Delhi-Iris & \textbf{93.75 $\pm$ 0.016} \\
ND-Iris-0405 & CASIA-Iris-Interval-v3 & \textbf{95.26 $\pm$ 0.009} \\
CASIA-Iris-Interval-v3 & IIT-Delhi-Iris & \textbf{93.86 $\pm$ 0.010} \\
CASIA-Iris-Interval-v3 & ND-Iris-0405 & \textbf{98.86 $\pm$ 0.002} \\
IIT-Delhi-Iris & ND-Iris-0405 & \textbf{98.92 $\pm$ 0.002} \\
IIT-Delhi-Iris & CASIA-Iris-Interval-v3 & \textbf{95.49 $\pm$ 0.008} \\
\hline
\multicolumn{3}{|c|}{\textbf{Baselines \cite{kerrigan2019iris}}} \\
\hline
\textbf{Method} & \textbf{Dataset} & \textbf{Accuracy} \\
\hline
OSIRIS & ND-Iris-0405 & 86.28 $\pm$ 6.50 \\
DRN & ND-Iris-0405 & 89.61 $\pm$ 5.08 \\
Context-100k & ND-Iris-0405 & 89.45 $\pm$ 3.85 \\
SegNet & ND-Iris-0405 & 89.75 $\pm$ 4.95 \\
\hline
\end{tabular}
\end{table}
\vspace{-2mm}
\vspace{-2mm}

\subsection{Cross-Dataset Model Generalization}
\begin{figure}[tb]
    \centering
    \begin{subfigure}[b]{0.48\textwidth}
        \includegraphics[width=\linewidth]{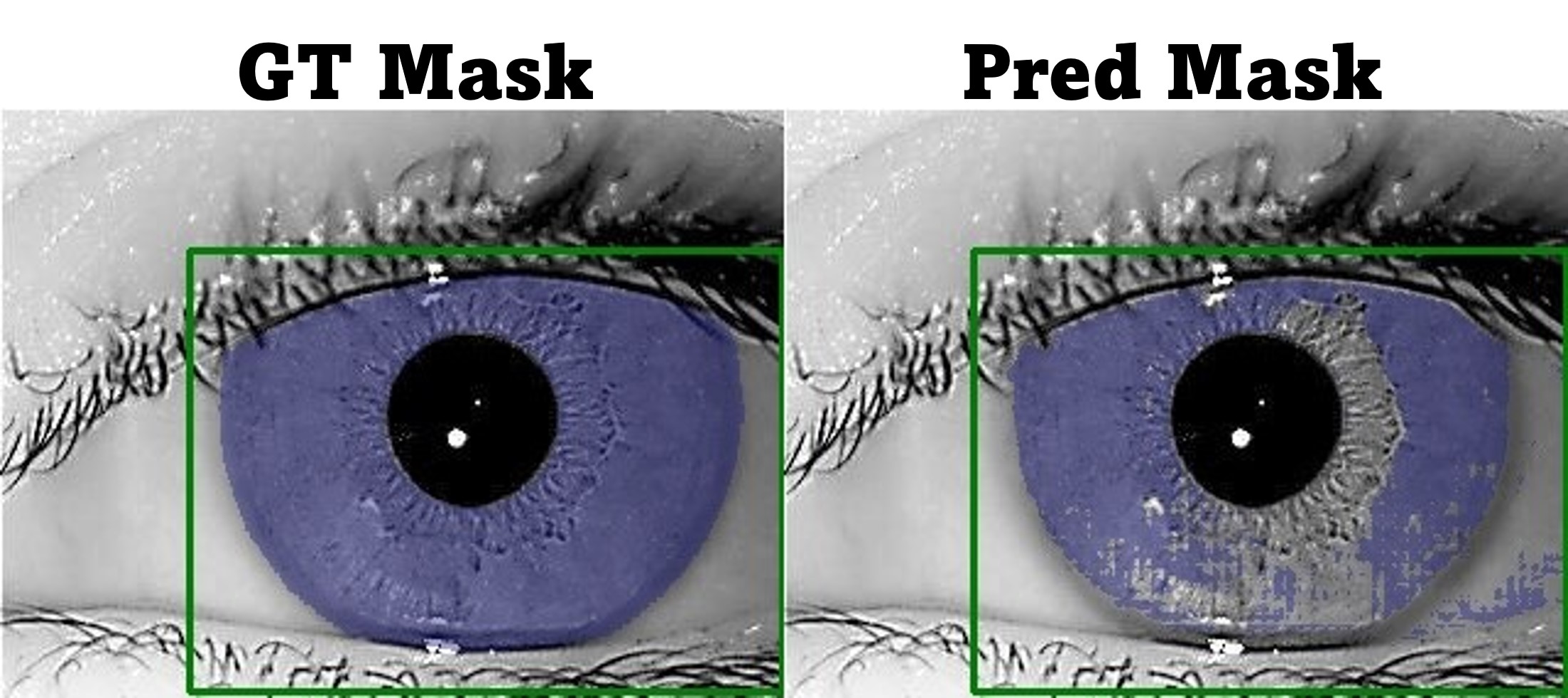}
        \caption{Identity 011\_01}
        \label{fig:011_01_n}  
    \end{subfigure}
    \hfill 
    \begin{subfigure}[b]{0.48\textwidth}
        \includegraphics[width=\linewidth]{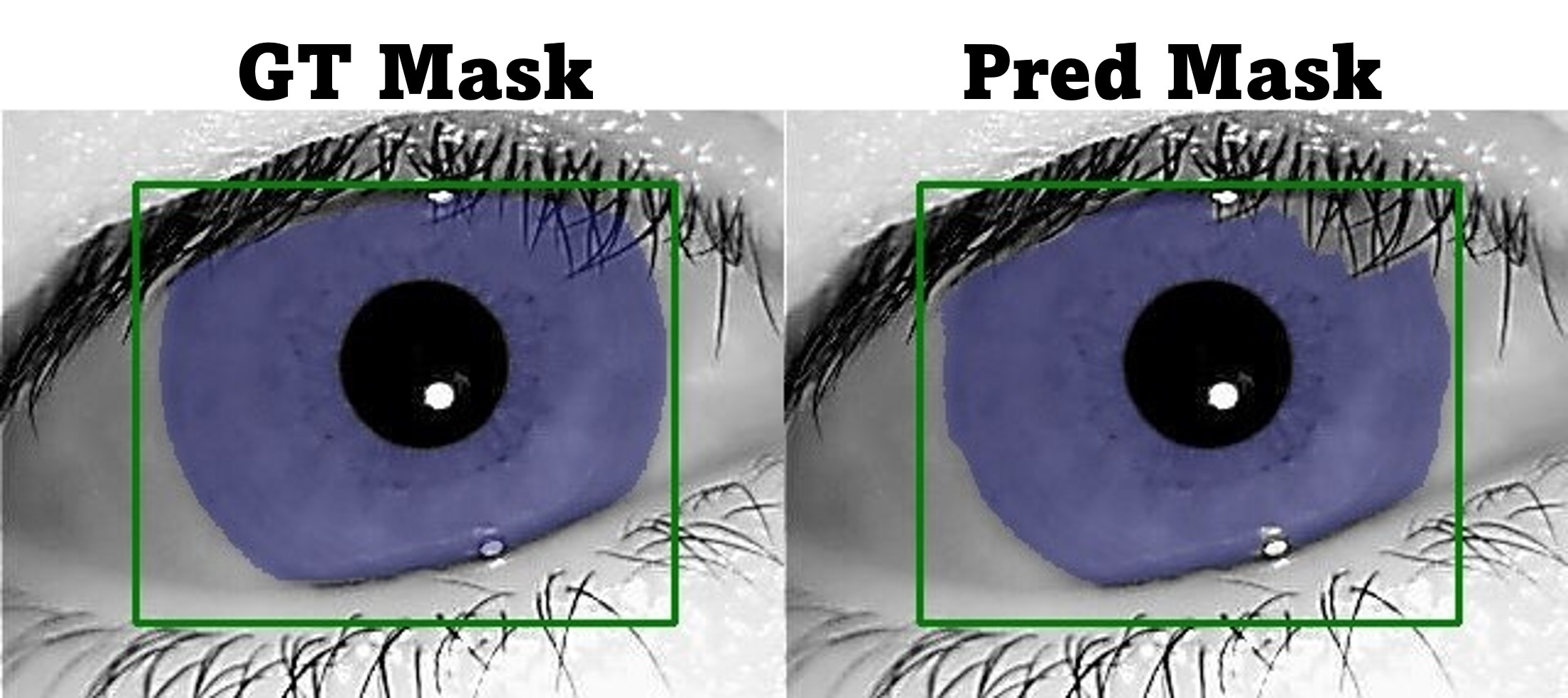}
        \caption{Identity 013\_06} 
        \label{fig:013_06_n}  
    \end{subfigure}
    \captionsetup{labelsep=period} 
  \captionsetup{labelfont=bf}
    \caption{Generalization results using ND-Iris-0405 (train) and IIT-Delhi-Iris (test). (a) A example of failure in segmentation. (b) An interesting observation of the model trying to segment the eyelashes on a previously unseen dataset.}
    \label{fig:generalization}  
\end{figure}
The generalization capability of our method was further put to the test by training on ND-Iris-0405 and evaluating on the IIT-Delhi-Iris and CASIA-Iris-Interval-v3 datasets resulting in a very competitive accuracy of 93.75\% and 95.26\%, respectively. For training, we utilized our fine-tuned model already trained on each dataset. This cross-dataset validation underscores our model's adaptability and its potential for deployment in varied scenarios (more generalization results are shown in \Cref{tab:average_iou}). 

\section{Is Iris-SAM More Accurate Than Ground Truth?}
\begin{figure}[ht]
    \centering
    \begin{subfigure}[b]{0.48\textwidth}
        \includegraphics[width=\linewidth]{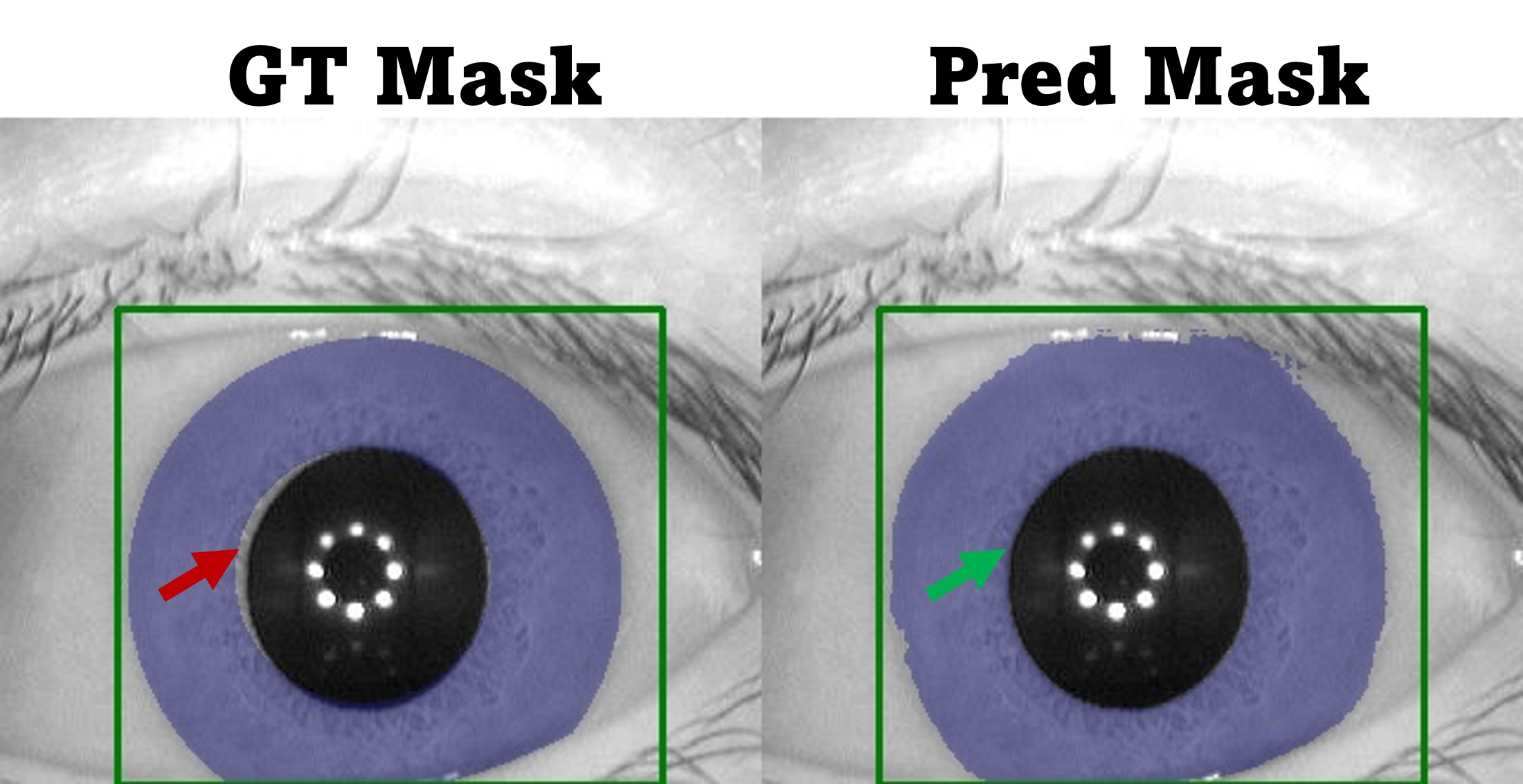}
        \caption{Identity S1019L04}
        \label{fig:S1019L04}
    \end{subfigure}
    \hfill
    \begin{subfigure}[b]{0.48\textwidth}
        \includegraphics[width=\linewidth]{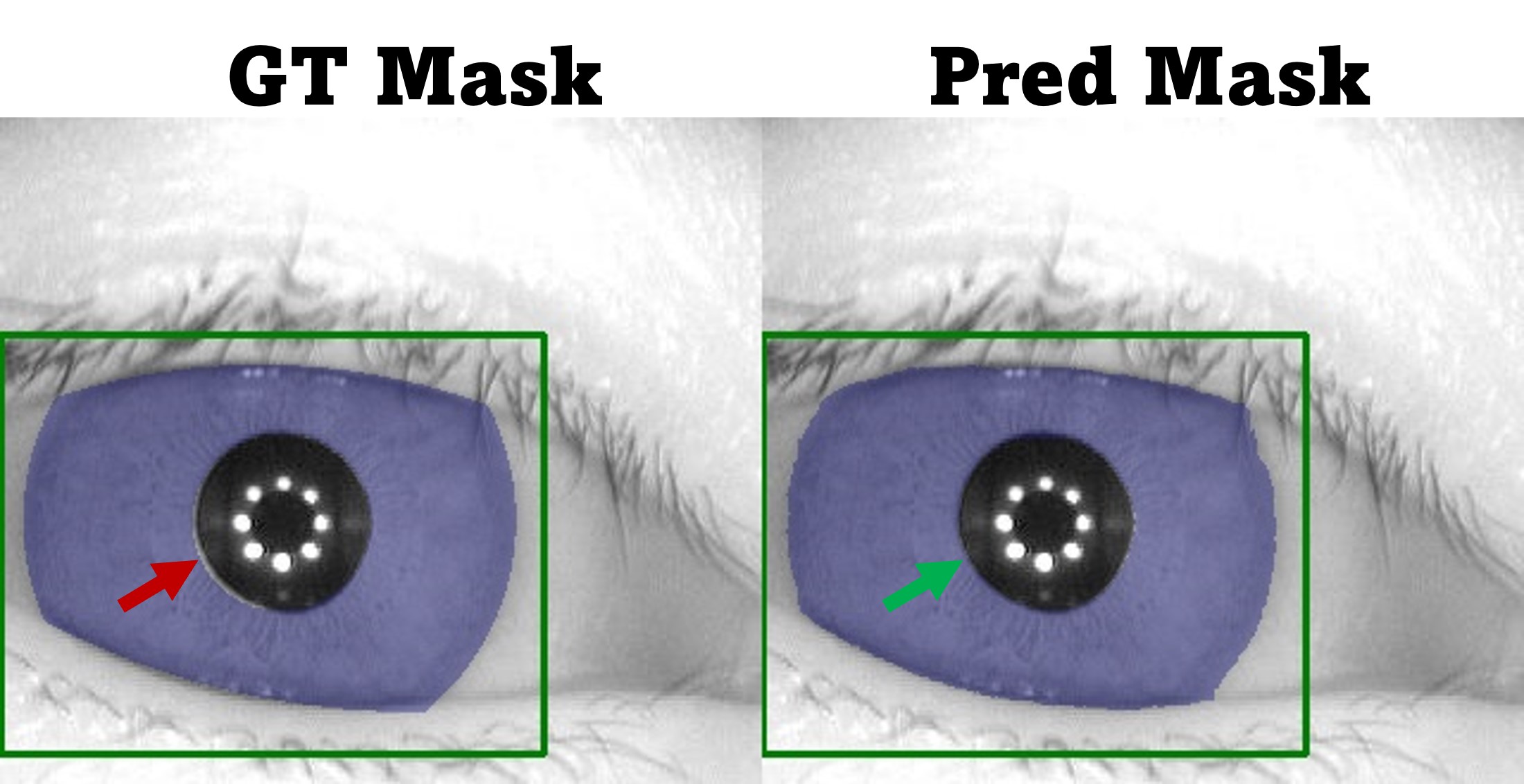}
        \caption{Identity S1009R05} 
        \label{fig:S1009R05}
    \end{subfigure}
      \captionsetup{labelsep=period} 
    \captionsetup{labelfont=bf}
    \caption{Examples where the predicted mask appears to be more accurate than the ground truth mask in the CASIA-Iris-Interval-v3 dataset.}
    \label{fig:bad-ground-truth1}
\end{figure}

\begin{figure}[ht]
    \centering
    \begin{subfigure}[b]{0.48\textwidth}
        \includegraphics[width=\linewidth, height=3.1cm]{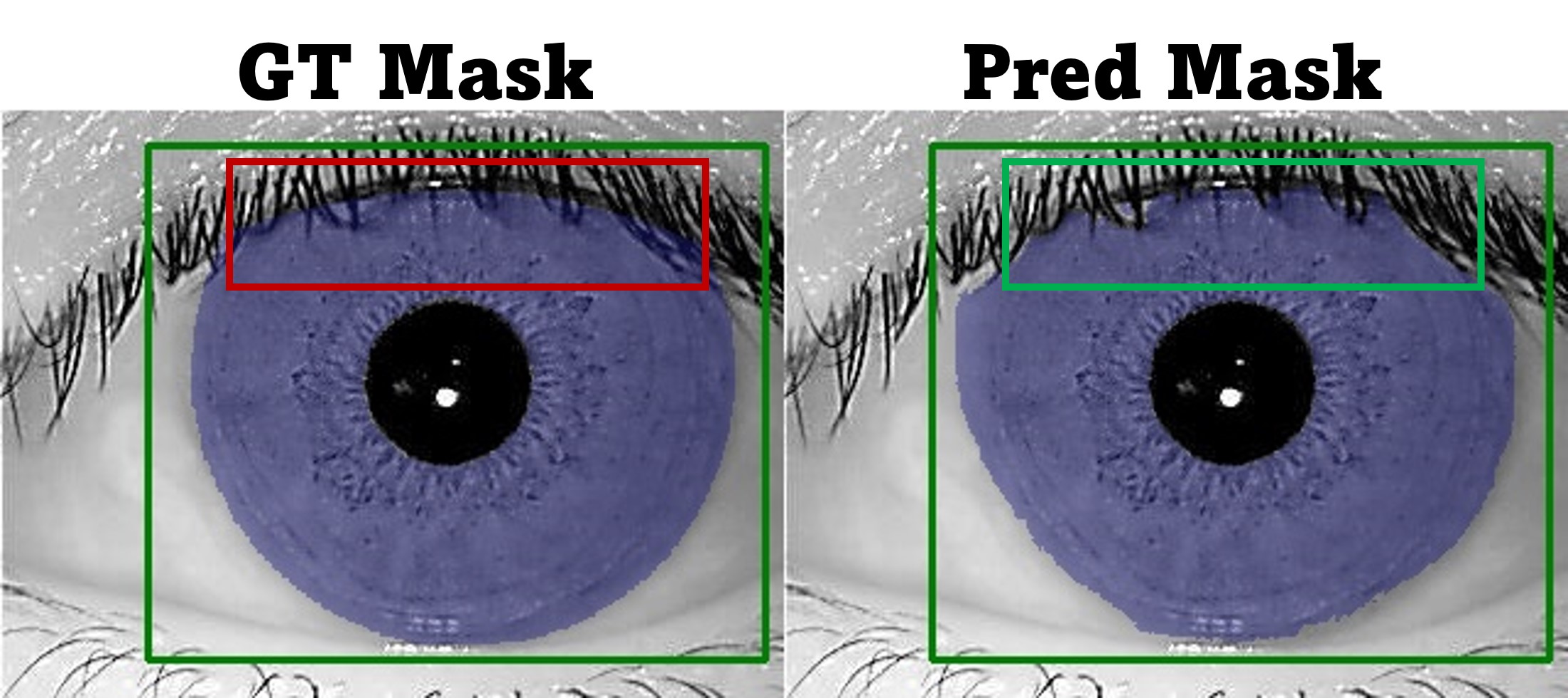}
        \caption{Identity 032\_07}
        \label{fig:032-07}
    \end{subfigure}
    \hfill
    \begin{subfigure}[b]{0.48\textwidth}
        \includegraphics[width=\linewidth, height=3.1cm]{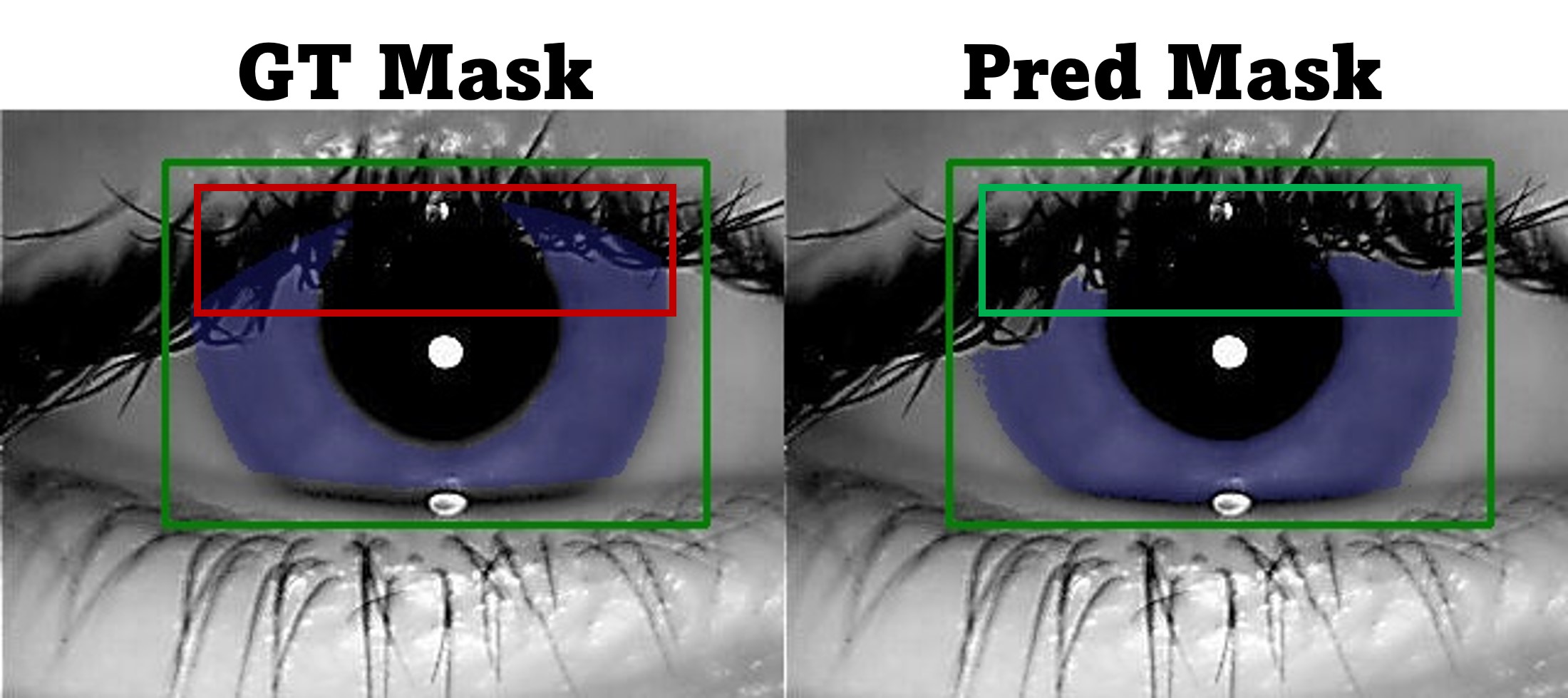}
        \caption{Identity 019\_05}
        \label{fig:019-05}
    \end{subfigure}
    \captionsetup{labelsep=period} 
    \captionsetup{labelfont=bf}
    \caption{Examples where the predicted mask appears to be more accurate than the ground truth mask in the IIT-Delhi-Iris dataset.}
    \label{fig:bad-ground-truth2}
\end{figure}

A notable observation from our experiments, as illustrated in \Cref{fig:bad-ground-truth1} and \Cref{fig:bad-ground-truth2}, is the occasional superiority of the model-predicted masks over the ground-truth (GT) masks.

\vspace{-3mm} 
In \Cref{fig:bad-ground-truth1}, marked by a red arrow, we observe an instance where the ground-truth mask partially omits a segment of the iris. In contrast, the model's prediction captures this missing section, demonstrating a nuanced understanding of the iris structure. This instance is especially interesting considering the model achieved this higher accuracy with a relatively straightforward fine-tuning process. It suggests that the model, guided by the Focal Loss, has developed a nuanced perception of iris features, surpassing even the manually annotated ground truth in certain aspects. \Cref{fig:bad-ground-truth2} presents a scenario where the ground-truth mask inaccurately depicts eyelashes as part of the iris. However, the model's predicted mask displays an interesting discernment by correctly excluding the eyelashes from the iris segmentation. This precision highlights the model's capability to distinguish between closely situated but distinct features within the ocular region, a crucial requirement for reliable iris segmentation. Firstly, the model's refined handling of ground-truth annotations showcases its deep understanding of iris morphology. Secondly, its precise discrimination between the iris and adjacent elements underscores the model's accuracy, which is essential for the reliability of iris biometric systems.
\begin{figure}[ht]
    \centering
    \begin{subfigure}[b]{0.24\textwidth}
        \includegraphics[width=\linewidth]{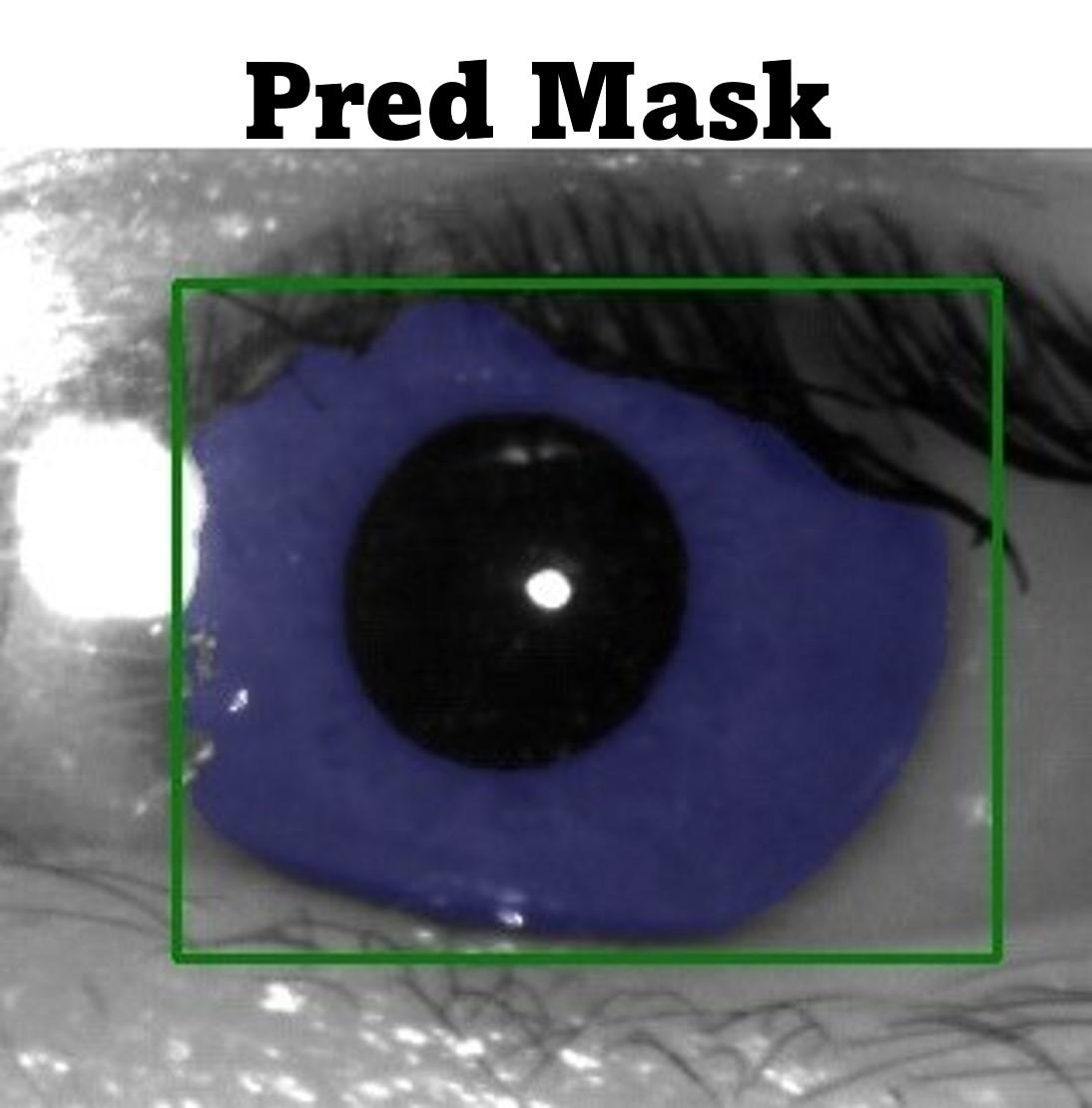}
        \label{fig:087-01}  
    \end{subfigure}
    \hfill
    \begin{subfigure}[b]{0.24\textwidth}
        \includegraphics[width=\linewidth]{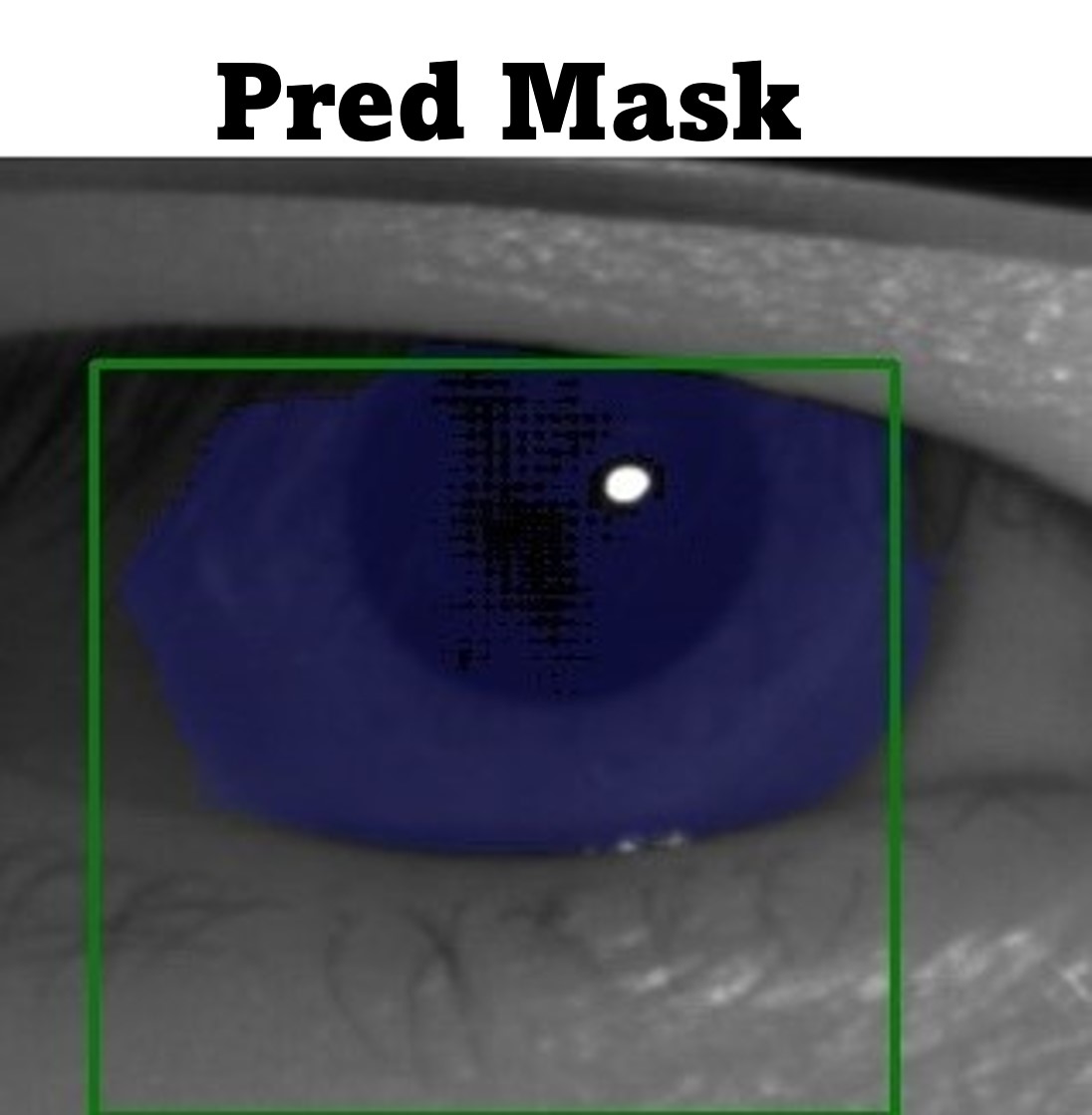}
        \label{fig:156-09}  
    \end{subfigure}
    \hfill
    \begin{subfigure}[b]{0.24\textwidth}
        \includegraphics[width=\linewidth]{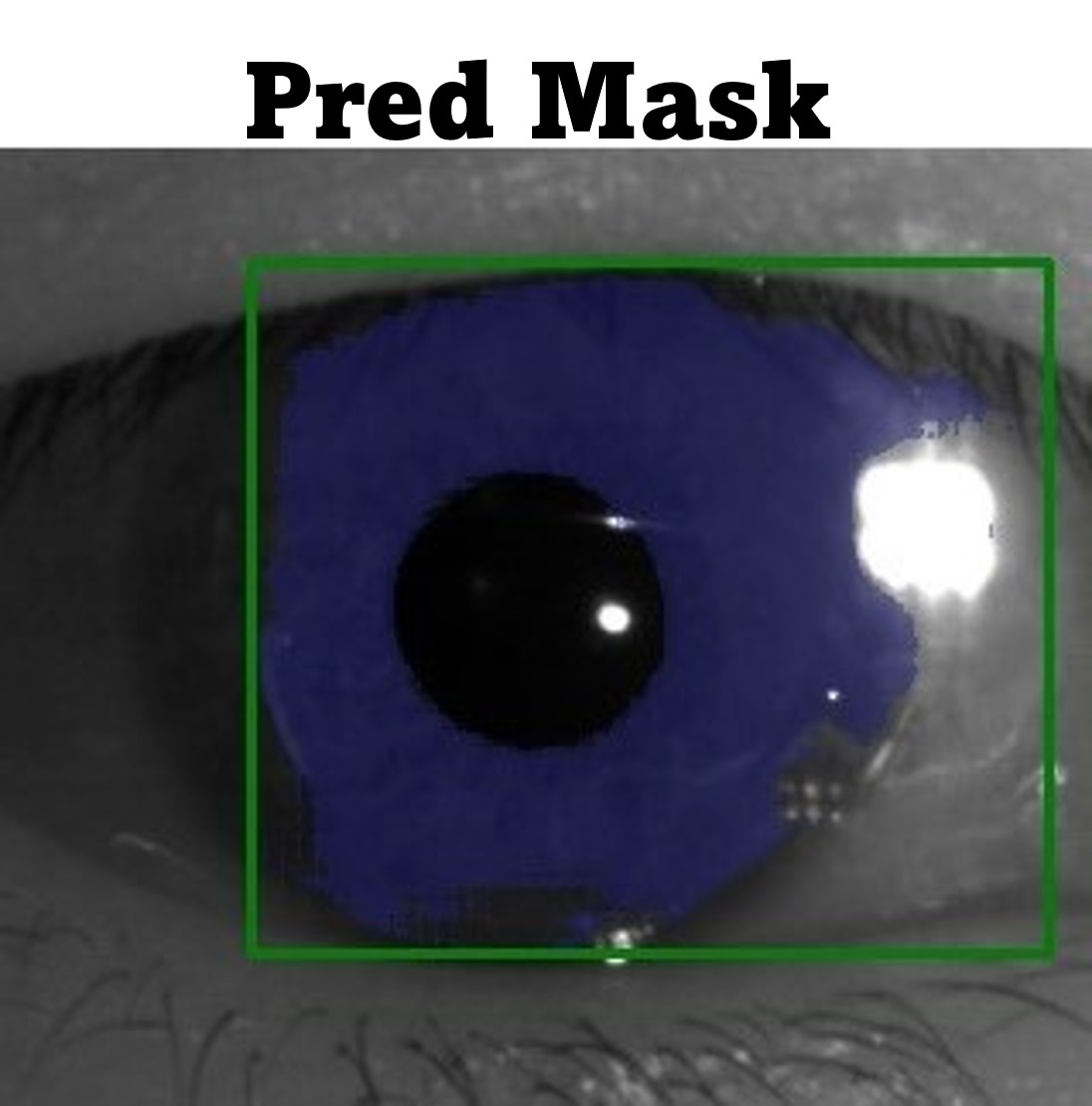}
        \label{fig:image3}  
    \end{subfigure}
    \hfill
    \begin{subfigure}[b]{0.24\textwidth}
        \includegraphics[width=\linewidth]{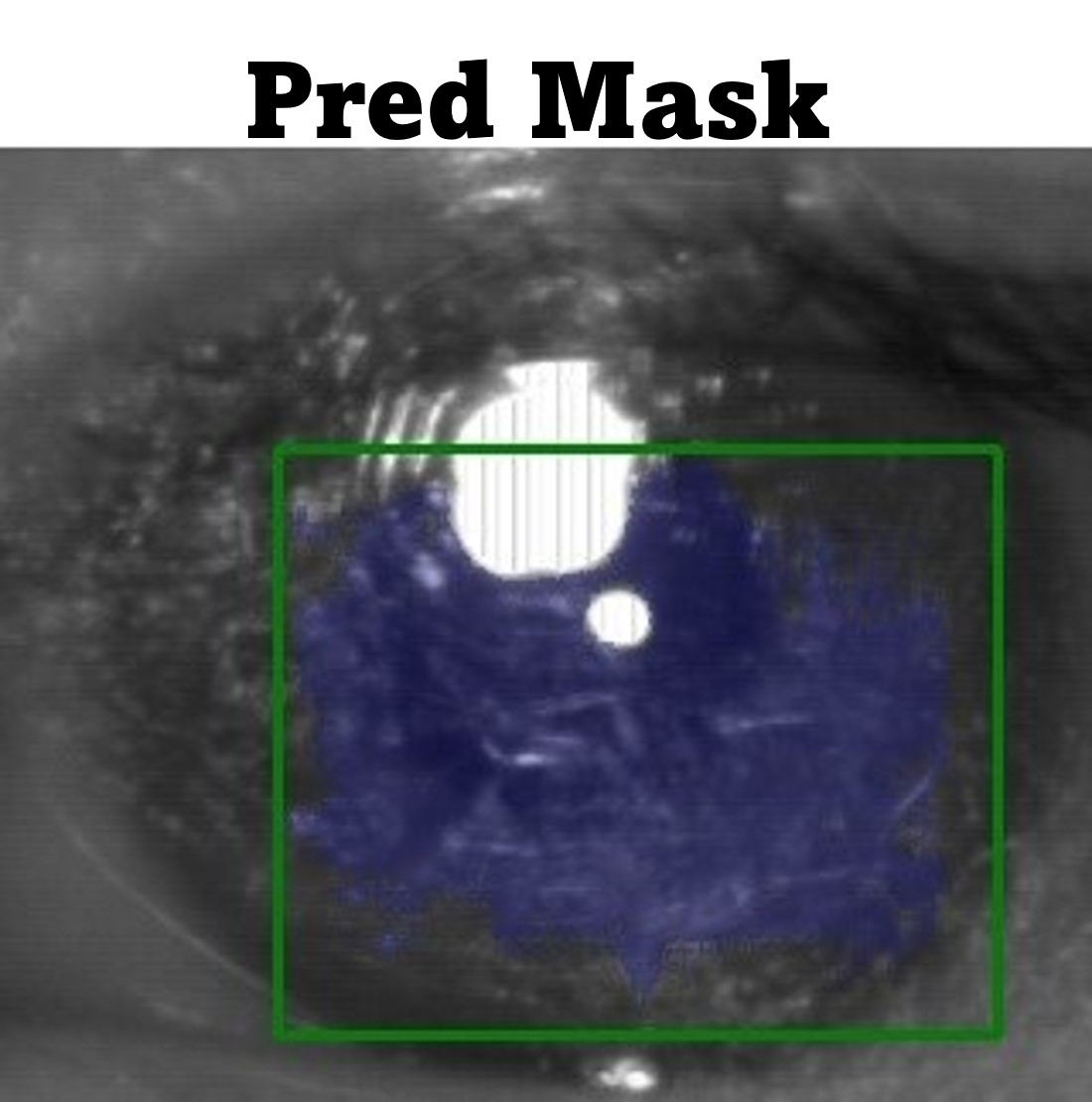}
        \label{fig:image4}  
    \end{subfigure}
    \captionsetup{labelsep=period} 
    \captionsetup{labelfont=bf}
    \caption{Evaluation of our model on a challenging dataset (irides with glasses) without further tuning. As can be seen, the results vary, with some instances showing good outcomes and others less favorable.}
    \label{fig:generalization-challange}  
\end{figure}
\begin{figure}[H]
    \centering
    \begin{subfigure}[b]{0.49\textwidth}
        \includegraphics[width=1.0\textwidth]{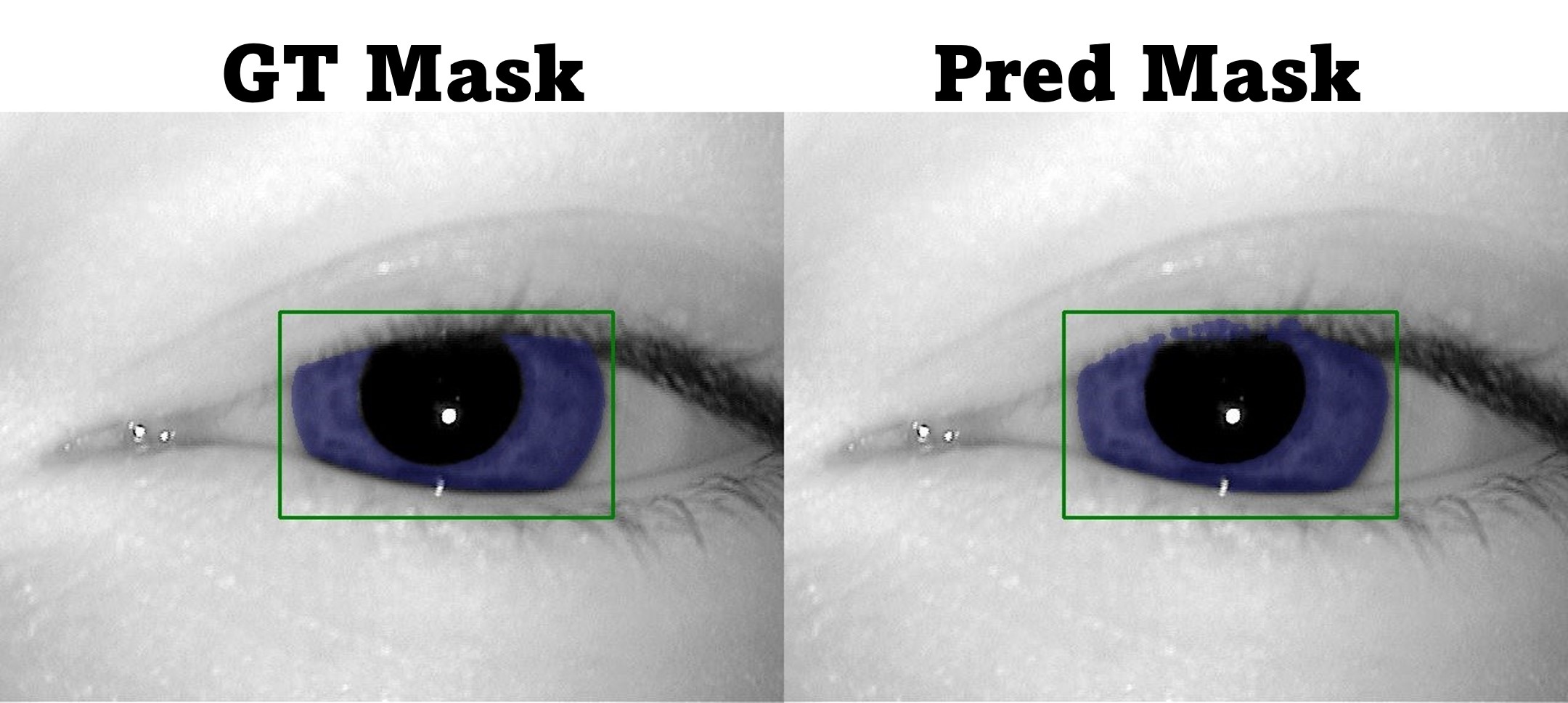}
        \caption{IoU = 98.23\%}
        \label{fig:04349d773}
    \end{subfigure}
\hfill
\begin{subfigure}[b]{0.5\textwidth}
        \includegraphics[width=1.0\textwidth]{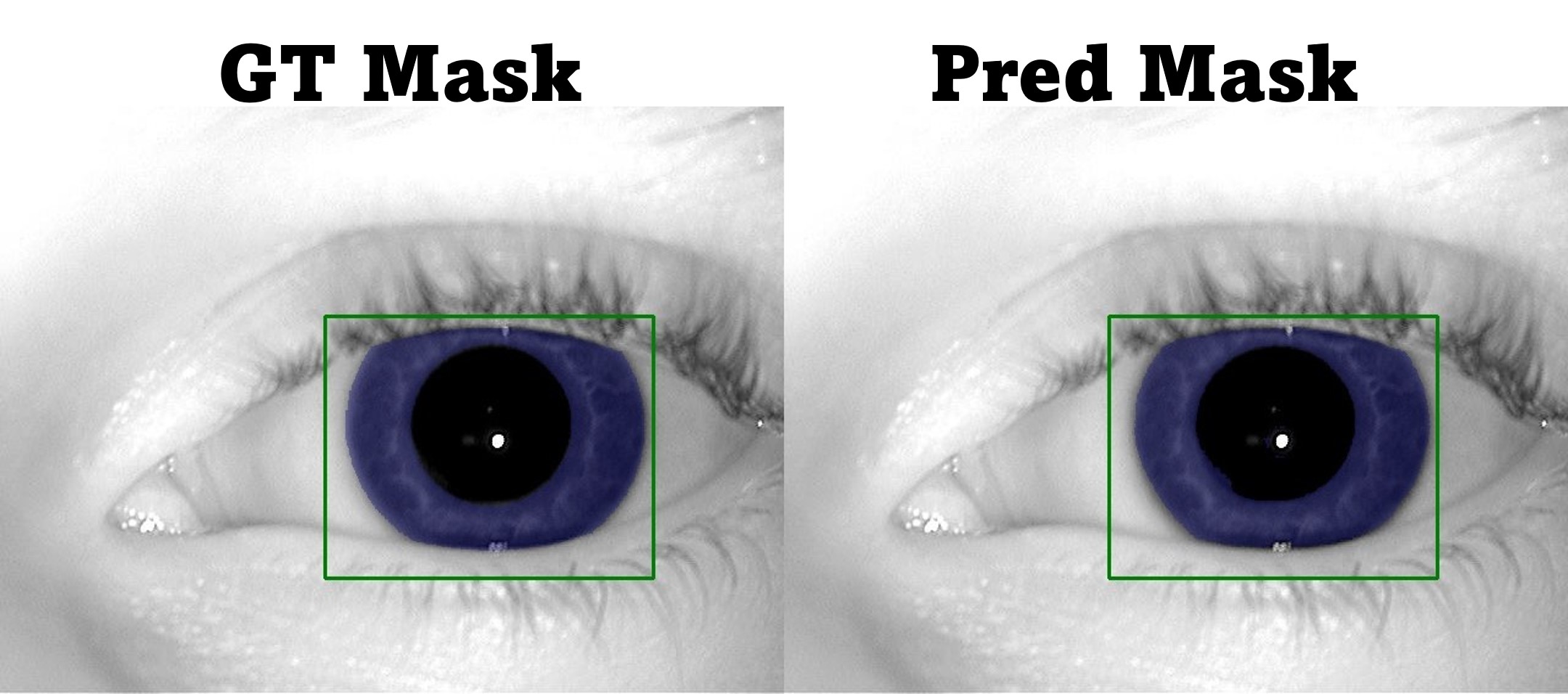}
        \caption{IoU = 97.70\%}
        \label{fig:04370d363}
    \end{subfigure}
    \captionsetup{labelsep=period} 
   \captionsetup{labelfont=bf}
    \caption{Samples from the ND-Iris-0405 dataset illustrating segmentation results with low IoU scores.}
    \label{fig:low-iou}
\end{figure}
To further examine our method and its robustness, we also tested it on a more challenging dataset. In the provided images depicting failure cases (\Cref{fig:generalization-challange}), it is apparent that challenges arise primarily from occlusions and reflections, common in practical scenarios such as eyewear (glasses) and varied lighting conditions. The presence of eyewear introduces additional reflective surfaces and distortions that complicate the segmentation process. Similarly, difficult lighting conditions can produce glares or shadows on the iris, further complicating accurate segmentation. These factors can lead to inaccuracies where the predicted mask either overextends beyond the actual iris boundaries or fails to encompass the entire iris, as the model struggles to differentiate between the iris, the occlusions, and reflections.
\section{Conclusion and Future Work}
In our study, the enhanced SAM model with Focal Loss has significantly advanced iris segmentation, achieving Average IoU scores of 96.94\% and 99.58\% on CASIA-Iris-Interval-v3 and ND-Iris-0405 datasets, respectively, and maintaining consistent performance with a very low standard deviation. Notably, the model handled complex scenarios, including dense eyelashes, as evidenced by its performance on the IIT-Delhi-Iris dataset, though it highlighted areas for future refinement. This work sets a new benchmark for iris segmentation and paves the way for the exploration of other challenges in iris segmentation, such as the processing of off-axis irides and images acquired in other spectral bands besides NIR.

\bigskip
\noindent \textbf{Acknowledgement.} We thank the University of Salzburg and Halmstad University for providing the ground truth datasets \cite{alonso2015near, hofbauer2014ground}. The code supporting the findings of this study is publicly available.\\\small(\href{https://github.com/ParisaFarmanifard/Iris-SAM}{Repository: https://github.com/ParisaFarmanifard/Iris-SAM})\small


\renewcommand{\bibname}{References}
\bibliographystyle{splncs04}
\bibliography{main}

\end{document}